\documentclass[sigconf]{acmart}
\pdfoutput=1
\renewcommand\footnotetextcopyrightpermission[1]{}

\usepackage{subfigure}
\usepackage{algorithmic}
\usepackage{multirow}
\usepackage{graphicx}
\usepackage{textcomp}
\usepackage{xcolor}
\usepackage{enumitem}
\usepackage{comment}
\usepackage{pifont}
\usepackage{amsthm}
\usepackage{algorithmic}
\usepackage{algorithm}
\usepackage{hyperref}
\usepackage[hyphenbreaks]{breakurl}
\newtheorem{definition}{Definition}

\AtBeginDocument{%
  \providecommand\BibTeX{{%
    \normalfont B\kern-0.5em{\scshape i\kern-0.25em b}\kern-0.8em\TeX}}}

\author{Yuan~Yuan}
\affiliation{
\institution{Department of Electronic Engineering \\ Tsinghua University \\ Beijing, China}
\country{}
}

\author{Huandong Wang}
\affiliation{
\institution{Department of Electronic Engineering \\ Tsinghua University \\ Beijing, China}
\country{}
}

\author{Jingtao Ding}
\affiliation{
\institution{Department of Electronic Engineering \\ Tsinghua University \\ Beijing, China}
\country{}
}

\author{Depeng Jin}
\affiliation{
\institution{Department of Electronic Engineering \\ Tsinghua University \\ Beijing, China}
\country{}
}

\author{Yong Li}
\affiliation{
\institution{Department of Electronic Engineering \\ Tsinghua University \\ Beijing, China}
\country{}
}

\setcopyright{rightsretained} 
\copyrightyear{2023} 
\acmYear{2023} 
\acmConference[WWW '23]{Proceedings of the ACM Web Conference
2023}{May 1--5, 2023}{Austin, TX, USA}
\acmBooktitle{Proceedings of the ACM Web Conference 2023 (WWW '23),
May 1--5, 2023, Austin, TX, USA}
\acmDOI{10.1145/3543507.3583276}
\acmISBN{978-1-4503-9416-1/23/04}

\begin{document}

\title[Learning to Simulate Daily Activities via  Modeling Dynamic Human Needs]{Learning to Simulate Daily Activities via  Modeling \\ Dynamic Human Needs}

\begin{abstract}
Daily activity data that records individuals' various types of activities in daily life are widely used in many applications such as activity scheduling, activity recommendation, and policymaking.
Though with high value, its accessibility is limited due to high collection costs and potential privacy issues. 
Therefore, simulating human activities to produce massive high-quality data is of great importance to benefit practical applications. 
However, existing solutions, including \textit{rule-based methods} with simplified assumptions of human behavior and \textit{data-driven methods} directly fitting real-world data, both cannot fully qualify for matching reality.  In this paper, motivated by the classic psychological theory, Maslow's need theory describing human motivation, we propose a knowledge-driven simulation framework based on generative adversarial imitation learning.
To enhance the fidelity and utility of the generated activity data, our core idea is to model the evolution of human needs as the underlying mechanism that drives activity generation in the simulation model.
Specifically, this is achieved by a hierarchical model structure that disentangles different need levels, and the use of neural stochastic differential equations that successfully captures piecewise-continuous characteristics of need dynamics.
Extensive experiments demonstrate that our framework outperforms the state-of-the-art baselines in terms of data fidelity and utility.
Besides, we present the insightful interpretability of the need modeling.  
The code is available at \url{https://github.com/tsinghua-fib-lab/SAND}.
\end{abstract}

\ccsdesc[500]{Computing methodologies}
\ccsdesc[500]{Computing methodologies~Modeling methodologies}
\ccsdesc[500]{Computing methodologies~Modeling and simulation}
\ccsdesc[500]{Computing methodologiFes~Model development and analysis}

\keywords{Daily activities, Simulation, Human needs, GAIL}

\maketitle

\begin{figure}[ht!]
    \centering
    \vspace{+2mm}
    \includegraphics[width=0.9\linewidth]{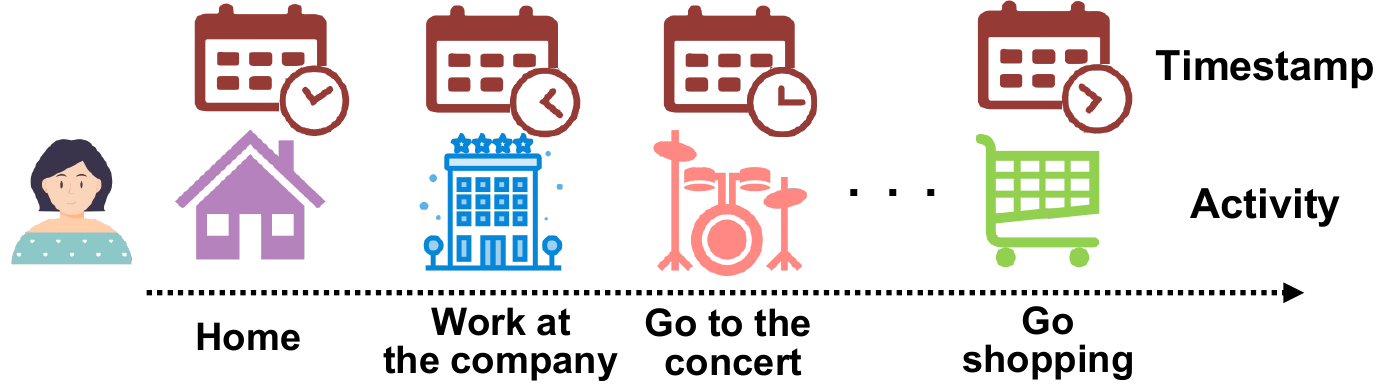}
    \caption{An example of activity sequences, where each entry contains information of the timestamp and activity type.}
    \label{fig:task}
    \vspace{-2mm}
\end{figure}

\section{introduction}
Web applications such as Yelp\footnote{https://www.yelp.com/} and Meituan\footnote{https://about.meituan.com/} have greatly improved the quality of people's daily life, and at the same time make it possible to record fine-grained activity data.
For example, as illustrated in Figure~\ref{fig:task}, daily life of an individual is usually logged as an activity sequence, \emph{i.e.}, $S=[a_1, a_2, ..., a_n]$, where each entry $a_i=(t_i, k_i)$ contains a timestamp $t_i\in\mathbb{R}^+$ and a discrete activity type $k_i \in \mathcal{C}$. 
Mining activity sequences is valuable for both research and industry in modeling user behaviors and supporting a wide range of applications, like activity planning and  recommendation~\citep{katz1983assessing, auld2012activity, wang2010toward}. 
Despite its high value, only a limited scale of such data is open-sourced for third-party researchers due to privacy-related restrictions on data sharing, which largely hinders the development of downstream applications~\citep{kim2019simulating,kim2020location}. 
Therefore, it is crucial to generate artificial data of human activities by simulation, which can reduce reliance on expensive real data and avoid privacy concerns.
In this paper, we study the problem of personalized user activity simulation that models individuals' decision process of what activity to perform at what time, and then generates artificial personalized activity data correspondingly. 
In order to be publicly shared and used as real-world data, the generated data is expected to be dissociated from real data, \emph{i.e.}, without privacy concerns, and meanwhile capable of retaining data fidelity and utility.

Existing solutions to this problem can be classified into two categories, \emph{i.e.}, \textit{rule-based methods} and \textit{data-driven methods}.
\textit{Rule-based methods} that simulate for activity scheduling~\citep{kitamura1996sequenced,ettema1993simulation,arentze2000albatross,bowman2001activity} have a basic assumption that activities can be described by predefined rules derived from activity theories such as utility maximization~\citep{recker1981toward}. 
However, real-world sequences exhibit complex transition patterns between activities with time dependence and high-order correlations, which are difficult to describe with prior simple rules~\cite{feng2020learning}.
Therefore, only relying on simplified assumptions makes \textit{rule-based methods} less qualified for modeling real-world activity behaviors. 
Instead,
\textit{data-driven methods} tackle this problem by directly fitting real-world data.
A series of sequential generative methods have been developed, from classical probability models, such as Markov models~\cite{prinz2011markov}, to deep learning models, such as Recurrent Neural Networks~(RNNs)~\cite{graves2013generating} and Generative Adversarial Imitation Learning~(GAIL)~\cite{ho2016generative}. 
Nevertheless, the above models cannot fully capture the temporal dynamics underlying human daily activities due to the unrealistic inductive bias of being time-invariant~\cite{rabiner1986introduction} or discrete updates only at observed time points~\cite{liang2021modeling}.
Comparatively, daily activities are always irregularly sampled and longer time intervals introduce larger uncertainty between observations, which requires a  deeper understanding and fine-grained characterization.

\begin{figure}[t]
    \centering
    \includegraphics[width=0.85\linewidth]{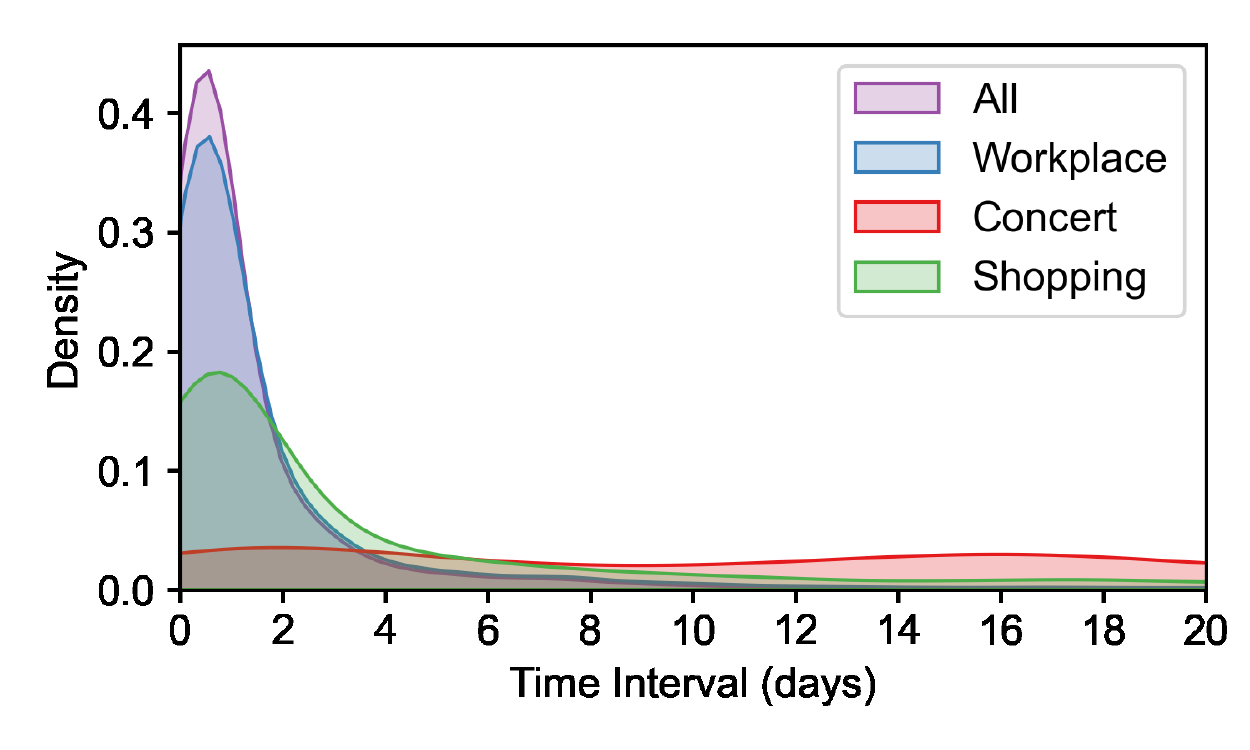}
    \caption{Interval distributions of different activities. Different activities inherently have distinct temporal dynamics.}
    \label{fig:task_analysis}
    \vspace{-3mm}
\end{figure}

More importantly, 
there exist complex and various patterns in terms of temporal dynamics of different activities, which are hard to discriminate from each other when mixed together.
For example, as Figure~\ref{fig:task_analysis} illustrates, time intervals of going to the ``Concert'' exhibit totally distinct patterns compared with going to the ``Workplace'' that is highly similar to ``All''. Although individuals lead generally regular daily routines, some activities still occur occasionally but cannot be ignored. However, with the overall distribution exhibiting long-tailed characteristics, the coarse-grained learning paradigm of state-of-the-art \textit{data-driven methods} can be easily biased by the uneven distribution and fail to adequately capture unique patterns of each activity. 
Therefore, to generate faithful data that matches reality, it is better not to solely rely on the observed data that may possibly reveal an overall but misleading activity pattern.

To address the above issues and achieve a realistic simulation, we propose a novel framework informed by psychological theories and integrate activity-related knowledge into the state-of-the-art GAIL method.  
Our key idea is to highlight the intrinsic drives of activity decisions, namely, \textbf{human needs},  which are well supported by Maslow's need theories.
Accordingly, human needs can be categorized into three levels: \textit{physiological needs}, \textit{safety needs}, and \textit{social needs}.
Guided by this knowledge, we explicitly model human needs in a data-driven manner.  We disentangle the needs behind daily activities to fully capture the aforementioned complex patterns in empirical data.
Specifically, we simultaneously model each need dynamics with an alternating process between
\textit{spontaneous flow} and \textit{instantaneous jump}. For example,  the accumulation of needs in evolution~(flow) triggers the occurrence of related activities while the decaying needs after satisfaction~(jump) can restrain tendencies towards specific activities.

In terms of the specific model design, the proposed GAIL-based framework consists of a discriminator that provides reward signals and a generator that learns to generate high-quality activities with a policy network.
Particularly, we utilize Maslow's Theory in our framework to enhance the activity simulation with need modeling from the following two perspectives.
First, to overcome the challenge of complex activity patterns, we design a hierarchical structure in the modeling to disentangle different need levels  and explicitly incorporate the underlying influence of human needs on activity decisions.
Second, to address the limitations of RNN-based methods in modeling continuous-time dynamics, we leverage Neural Stochastic Differential Equations~\citep{jia2019neural} to capture piecewise-continuous characteristics of need dynamics alternating between \textit{spontaneous flow} and \textit{instantaneous jump}.
The above need dynamics further serve as the states that define the policy function, which calculates activity intensities based on the current need state and decides the next action accordingly.
In conclusion, our contributions can be summarized as follows:
\begin{itemize}[itemsep=2pt,topsep=0pt,parsep=0pt,leftmargin=*]
    \item We are the first to explicitly model the intrinsic drives of activities, \emph{i.e.},  human needs,  which brings the synergy of psychological theories and data-driven learning.
    \item  We propose a novel knowledge-driven activity simulation framework based on GAIL, leveraging Maslow's theory to enhance the simulation reality by capturing need dynamics.
    \item Extensive experiments on two real-world datasets show the effectiveness of  the  framework in generating synthetic data regarding fidelity, utility, and interpretability.
\end{itemize}

\section{Preliminaries}

\textbf{Problem Statement.} Daily activity data can be defined as a temporal sequence of events $S=[a_1, a_2, ..., a_n]$, where $a_i$ is a tuple $(t_i,k_i)$, $t_i$ denotes the timestamp and $k_i$ is the activity type, \emph{e.g.}, eating at restaurants, working at companies, playing at sports centers. The problem of activity simulation can be defined as follows:

\begin{definition}[Human Activity Simulation]
Given a real-world activity dataset, generate a realistic activity sequence $\hat{S}=[\hat{a}_1, \hat{a}_2, ..., \hat{a}_n]$ with a parameterized generative model.
\end{definition}

\textbf{Temporal Point Process.} A temporal point process~(TPP)~\cite{mei2016neural} can be realized by an event sequence $\mathcal{H}_T=\{(t_1, k_1), ..., (t_n, k_n) | t_n < T\}$.
Here $t_i$ represents the arrival time of the event and $k_i$ is the event mark.  
Let $\mathcal{H}_t$ denote the history of past events up to time $t$, the conditional intensity function $\lambda^*_k(t)$ (the $k_{th}$ event category) is defined as: $\lambda^*_k(t) =\lim_{\Delta t \to 0^+}\frac{\mathbb{P}(\mathrm{event\ of\ type } k  \mathrm{\ in} [t, t+\Delta t]|\mathcal{H}_t)}{\Delta t}$. Note that $\lambda^*(t)=\sum\lambda^*_k(t)$ denotes the total conditional intensity, deciding the arrival time without considering event types. Then the event type is sampled at the probability proportional to $\lambda^*_k(t)$.

\textbf{Neural Ordinary Differential Equations.} NODE~\cite{chen2018neural} describes the evolution of the system state over continuous time $t\in\mathbb{R}^+$ by modeling the first-order ordinary differential equations with neural networks. Specifically, the derivative of the latent state is modeled as:  $d\mathbf{h}(t)=f(\mathbf{h}(t), t; \theta)\cdot dt$,  where $\mathbf{h}(t)$ is the latent state  and  $f$ parameterized by a neural network  describes the derivative at time $t$. The system output at time $t_1$ can be solved with an initial value at time $t_0$ by an ODE solver:  $\mathbf{h}(t_1) = \mathbf{h}(t_0)+\int_{t_0}^{t_1}f(\mathbf{h}(t), t; \theta)\cdot dt$.

In this work, we take the first attempt to characterize human needs with neural differential equations. 

\section{Method}

We first introduce how we model human needs to motivate the framework design in Section~\ref{sub:need_aware}, then explain the MDP modeling of the decision process in Section~\ref{sub:mdp}, and finally elaborate on the framework details in Section~\ref{sec:network_archi}. 

\subsection{Human Needs Modeling}\label{sub:need_aware}

\noindent\textbf{Hierarchy of Needs.}\label{sec:hierarchy} According to a classic theory in psychology, i.e., Maslow's Theory~\citep{maslow1943theory}, people are motivated to achieve a hierarchy of needs, including \textit{physiological needs}, \textit{safety needs}, \textit{social needs}, \textit{esteem needs}, and \textit{self-actualization needs}, in a priority order, where higher levels of need are modeled as long-term changes such as life stages. With the development of Maslow's Theory, the follow-up theories~\citep{chung1969markov,rauschenberger1980test,chang2008synthesized} have introduced flexibility in the hierarchy. For example, different needs can be pursued simultaneously, and there exist transition probabilities between any pair of needs.
We do not take the top two need levels for \textit{esteem} and \textit{self-actualization} into consideration because they are too abstract and their effects can only be observed in a long term. 

Here we classify individuals' activities into three need levels, including \textit{physiological needs} (level-1), \textit{safety needs} (level-2), and \textit{social needs} (level-3), which are sufficient to depict patterns of daily life~\citep{kim2020location,kavak2019location}. These three need levels are often triggered or satisfied in a short period, which are consistent with daily activities that happen within a short term (a few hours). 
We provide descriptions of each need level as follows:

\begin{itemize}[itemsep=2pt,topsep=0pt,parsep=0pt,leftmargin=*]
    \item \textbf{\textit{Physiological needs}} refer to biological requirements for survival, \emph{e.g.}, food, drink, and shelter. The human body cannot function optimally without satisfying these needs. 
    \item \textbf{\textit{Safety needs}} refer to requirements for security and safety, \emph{e.g.}, education and employment. Besides physiological needs, people expect their lives to be orderly, regular, and controllable. 
    \item \textbf{\textit{Social needs}} refer to requirements for spirits, \emph{e.g.}, entertainment and social relationships. After meeting physiological and safety needs, people are also striving for spiritual satisfaction.
\end{itemize}

In our modeling,  we follow Maslow's Theory in a more flexible way, rather than the original needs pursued in a rigid order. The fulfillment  order can be flexible according to individual preferences and external circumstances. 
Based on well-respected need theories, each activity is explicitly labeled with one of the need levels\footnote{We refer the readers to Section~\ref{sec:annotation} for more details of the need annotation.}. The association between human needs and activities based on expert knowledge bridges the gap between classic psychological theories and human behavior modeling, which provides opportunities to model human needs computationally in a data-driven manner.

\textbf{Evolution of Needs.}\label{sec:evolution} In real-world scenarios, human needs are not static but generally evolve with time dynamically, which not only derive from spontaneous changes, but also can be interrupted by happened activities. 
To better learn sequential activity patterns, it is essential to capture the underlying mechanism of need dynamics. 
However, it is non-trivial because human needs cannot be observed explicitly and are affected by various factors, such as activity relations and periodicity. Besides, different from activities that happen one by one, need dynamics are more complicated with synchronicity and competitiveness among different levels. 

To effectively capture the underlying need dynamics, we innovatively capture piecewise-continuous dynamics in human needs including  \textit{spontaneous flow} and \textit{instantaneous jump} as follows:

\begin{figure}[t]
    \centering
    \includegraphics[width=0.99\linewidth]{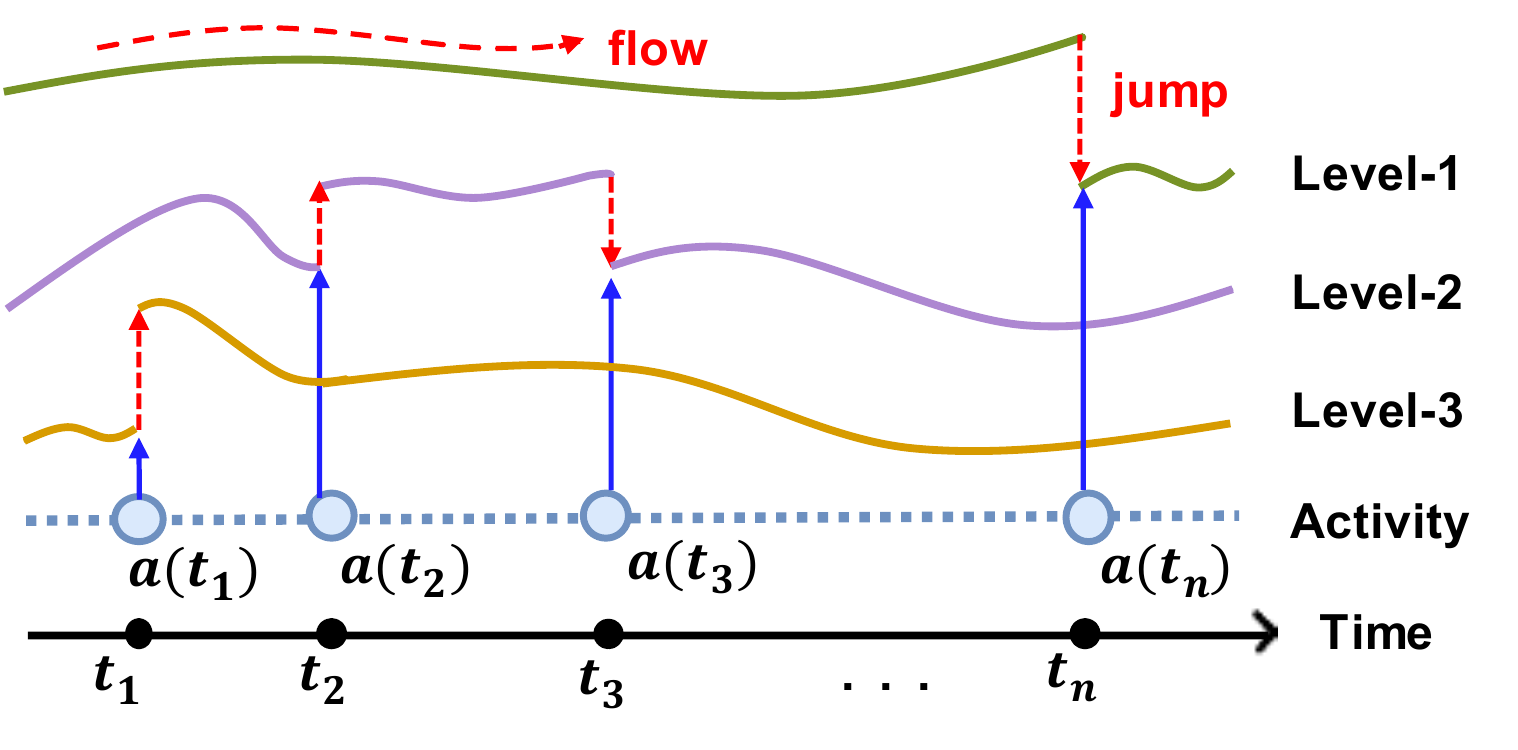}
    \caption{Illustration of the need evolution. Representations of three-level need states evolve continuously over time until interrupted  by a corresponding activity (\emph{e.g.}, $a(t_1)$ corresponds to level-1). Note that the need state is modeled by an embedding rather than a scalar, thus the jump up and down do not indicate an increase or decrease.}
    \label{fig:evolve}
\end{figure}

\begin{itemize}[itemsep=2pt,topsep=0pt,parsep=0pt,leftmargin=*]
    \item \textbf{\textit{Spontaneous flow}} denotes the continuous-time flow of need states. For example, needs for some activities can accumulate without taking them for a long time. Meanwhile, needs can also decay gradually as time goes by. 
    \item \textbf{\textit{Instantaneous jump}} models the influence of activities on the need states. For instance, the happened activities can immediately change the evolution trajectory of the corresponding need state.
\end{itemize}

\noindent Naturally, the two kinds of dynamics describe an active process of need evolution and need satisfaction.

Particularly, the three levels are disentangled in dynamic modeling, so they follow distinct evolution laws. Figure~\ref{fig:evolve} illustrates the two evolution mechanisms of different need levels. Nevertheless, it is challenging to learn such dynamics since needs are intrinsically unobserved and stochastic with the coexistence of continuity and jump. To tackle this problem, we represent human needs with a stochastic embedding process $\mathbf{z}(t)$ defined as follows:

\begin{definition}[Need Embedding Process]
The need embedding processes are $\{\mathbf{z}_i(t), i\in \{1,2,3\}, t\geq 0\}$, where $\mathbf{z}_i(t)$ is the representation of the $i_{th}$ need level at time $t$.
\end{definition}

\noindent In the above definition, we depict human needs with an embedding process $\mathbf{z}(t)$ instead of a direct scalar value for stronger representation capabilities. Particularly, $\mathbf{z}(t)$ is composed of three components $\mathbf{z}_1(t)$, $\mathbf{z}_2(t)$, $\mathbf{z}_3(t)$ that correspond to different need levels. Then the need embedding process $\mathbf{z}(t)$ with both \textit{spontaneous flow} and \textit{instantaneous jump} can be formulated as follows:

\begin{small}
\begin{equation}
   \left\{
   		\begin{array}{lr}
   		\mathbf{z}(t+dt)=\mathbf{z}(t)+\mathcal{F}(t, \mathbf{z}(t))dt, \mathrm{\ \ no\  activity\  in\  } [t,t+dt), \\
   		  \\
   		\lim\limits_{\Delta t \to 0^+}\mathbf{z}(t_i+\Delta t)=\mathcal{G}(t_i, \mathbf{z}(t_i),k(t_i)), \mathrm{\ \ with\  activity\ } k\  \mathrm{at\ the\  time\ } t_i, 
   		\end{array}
   \right.\label{eq:z_F_G}
\end{equation}
\end{small}

\noindent where $\mathcal{F}$ and $\mathcal{G}$\footnote{The time dependent variables in our modeling are all left continuous in $t$, \emph{i.e.}, $\lim\limits_{\epsilon \to 0^+}\mathbf{z}(t-\epsilon)=\mathbf{z}(t)$. } control the \textit{spontaneous flow} and \textit{instantaneous jump}, respectively, and $k(t_i)$ denotes the the occurred activity.

\subsection{Sequential Decision Processes}\label{sub:mdp}

The generation of activity sequences depends on individuals' decisions on what activity to take based on his/her own need state step by step. The whole process consists of a sequence of activity decisions that aim to maximize the total received "reward" along the process. Here we model the decision process as a Markov decision process (MDP)~\cite{sutton1998introduction}, and it is described by a 4-tuple $<\mathcal{S, A, T, R}>$, where $\mathcal{S}$ is the state space, $\mathcal{A}$ is the action space, $\mathcal{T}$ is the state transition, and $\mathcal{R}$ is reward function. The basic elements of MDPs are : (i) \textbf{State} represents the current need state. (ii) \textbf{Action} is generated based on the state by sampling a time interval $\tau$ and an activity type $k$. (iii) \textbf{Policy function} decides the next activity time and type.  (iv)  \textbf{State transition} controls how the state updates with two transit laws, \emph{i.e.}, spontaneous flow and instantaneous jump. (v) \textbf{Reward function} evaluates the utility of taking the action under the state, which is unknown and has to be learned from the data.

Given the activity history $\mathbf{s}_t=\{(t_i,k_i)\}_{t_i<t}$, the stochastic policy function $\pi_\theta(a|\mathbf{s}_t)$  samples an interval time $\tau$ and an activity type $k$ to generate the next activity $a=(t_{i+1},k_{i+1})$, where $t_{i+1}=t_i+\tau$. Then, a reward value is calculated and the state will be updated by an \textit{instantaneous jump}. 
Besides, there are also feedbacks of need states to the individual over time known as the \textit{spontaneous flow}.

\subsection{Proposed Framework: SAND}\label{sec:network_archi}

\begin{figure}[t]
    \centering
    \includegraphics[width=1\linewidth]{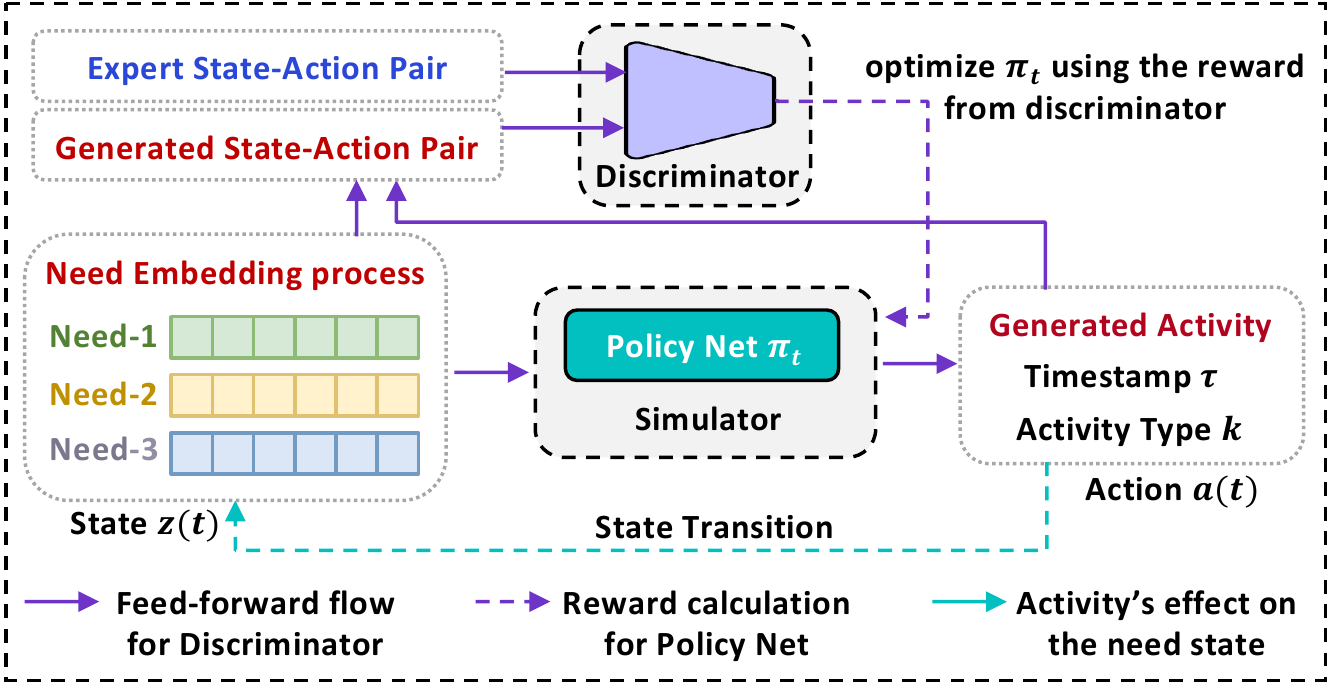}
    \caption{Illustration of the SAND framework. The policy and discriminator networks are optimized adversarially, and the state transition consists of two evolution mechanisms.}
    \label{fig:framework}
    \vspace{-4mm}
\end{figure}

In this section, we present a novel framework, SAND, which \underline{\textbf{S}}imulates human \underline{\textbf{A}}ctivities with \underline{\textbf{N}}eed \underline{\textbf{D}}ynamics.
Overall, it provides the synergy of need theories and imitation learning in simulating the activity decision-making process.
As shown in Figure~\ref{fig:framework}, it learns the policy and reward functions adversarially, where the need embedding process $\mathbf{z}(t)$ plays an essential role in the loop. We elaborate on the details of key components in the following sections.

\subsubsection{\textbf{Learning Need Dynamics.}}  To model the need dynamics including the \textit{spontaneous flow} and \textit{instantaneous jump}, we utilize neural stochastic differential equations~\cite{jia2019neural} to describe such continuity and discontinuity, where the need embedding process $\{\mathbf{z}_i(t), i\in \{1,2,3\}, t\geq 0\}$ acts as the latent state. 
Between activity observations, each $\mathbf{z}_i(t)$ flows continuously over time. Once an activity happens, the corresponding need embedding process is interrupted by a state jump. Different from directly modeling the changes of the hidden state like RNNs~\cite{wu2017modeling}, neural differential equations model the derivative of  $\mathbf{z}(t)$ to better capture the continuous-time characteristics.
Specifically, the derivative of the $i_{th}$ need state is formulated as follows:

\begin{small}
\begin{equation}
    d\mathbf{z}_i(t) = f_i(\mathbf{z}_i(t),t;\theta_i)\cdot dt + \omega_i(\mathbf{z}_i(t),\mathbf{k}_i(t),t;\gamma_i)\cdot dN_i(t) \ ,\label{eq:derivative}
\end{equation}
\end{small}

\noindent where $f_i$ and $\omega_i$ are both parameterized by neural networks and  control the \textit{spontaneous flow} and \textit{instantaneous jump} of the $i_{th}$ need embedding process, respectively, and $N_i(t)$ records the number of activities of the $i_{th}$ level up to time $t$. $f$ and $\omega$ in Eq.~\eqref{eq:derivative} are implementations of the function $\mathcal{F}$ and $\mathcal{G}$ defined in Eq.~\eqref{eq:z_F_G}. 
In particular, each state $\mathbf{z}_i(t) \in \mathbb{R}^n$ is composed of two vectors: (1) $\mathbf{c}_i(t) \in \mathbb{R}^{n_1}$ encodes the internal need state, and (2) $\mathbf{h}_i(t) \in \mathbb{R}^{n_2}$ encodes effects of the historical activities.

\begin{figure}[t]
    \centering
    \subfigure[Spontaneous flow]{
        \label{fig:dynamic_net_flow}
        \includegraphics[width=0.99\columnwidth]{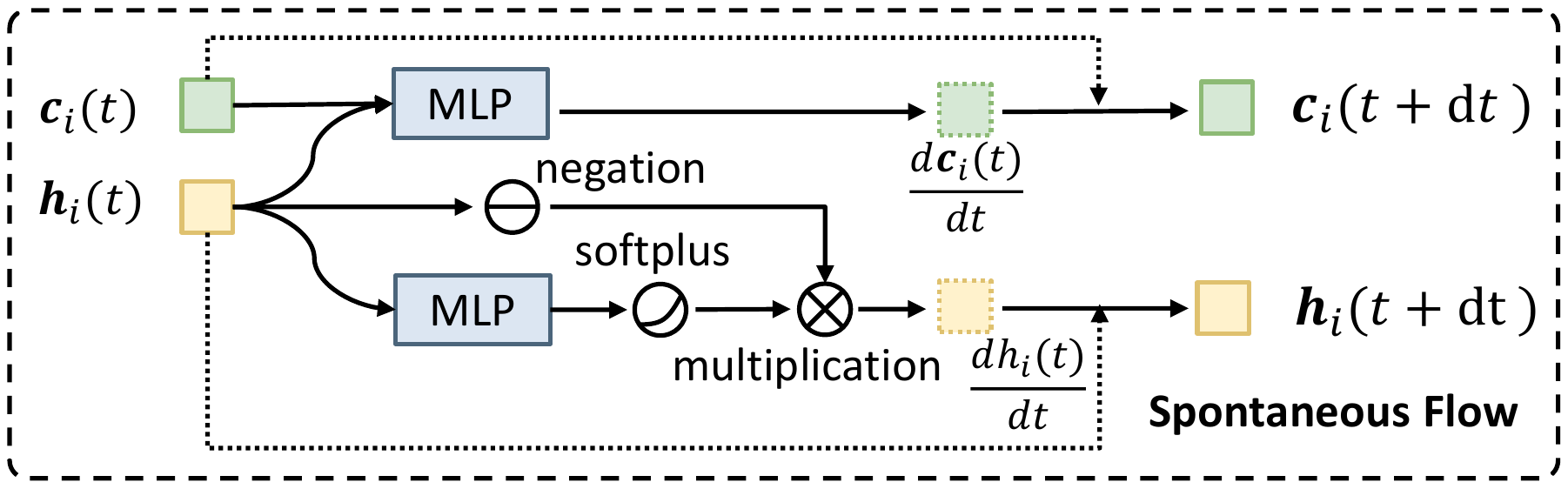}}
    \subfigure[Instantaneous jump]{
        \label{fig:dynamic_net_jump}
        \includegraphics[width=0.99\columnwidth]{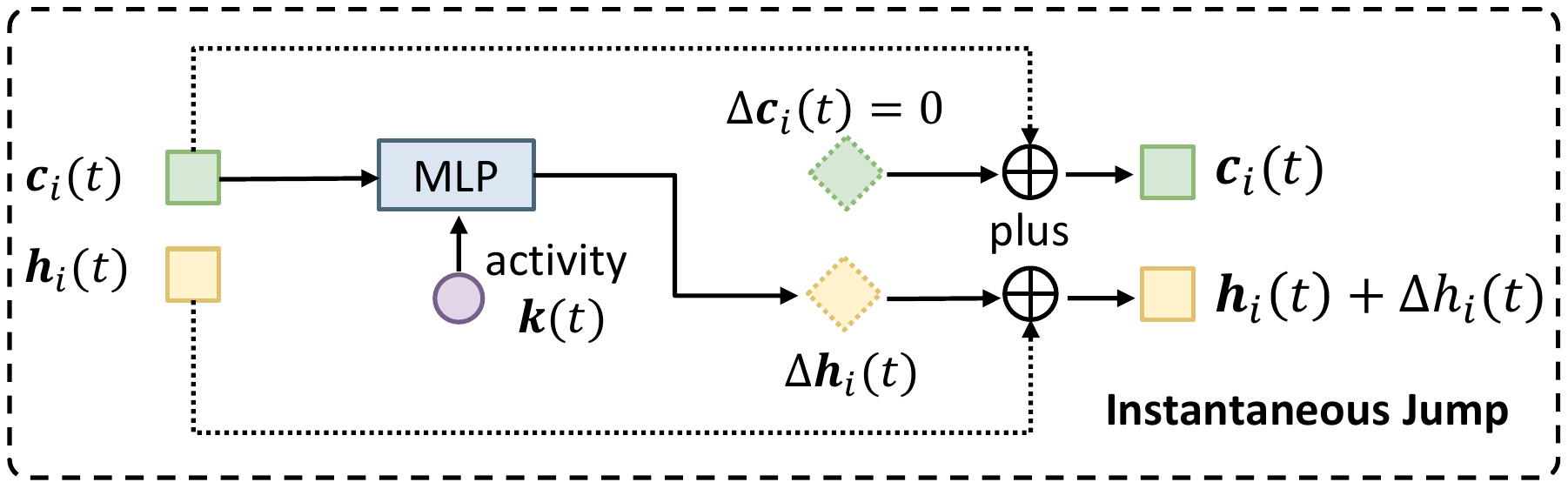}}
    \caption{Network architecture to learn need dynamics with both spontaneous flow and instantaneous jump. (a) shows the spontaneous flow of $\mathbf{c}_i(t)$ and $\mathbf{h}_i(t)$ based on the derivatives $\frac{d\mathbf{c}_i(t)}{dt}$ and $\frac{d\mathbf{h}_i(t)}{dt}$. (b) illustrates the instantaneous jump caused by the happened activity $k$.}
    \label{fig:dynamic_net}
\end{figure}

\textbf{\textit{Spontaneous flow.}}
The top part in Figure~\ref{fig:dynamic_net} shows the network design to model spontaneous flow. The neural function $f_i$ in Eq.~\eqref{eq:derivative} controls the spontaneous flow of the state $\mathbf{z}_i(t)$. 
Although $\mathbf{z}_i(t)$ contains two vectors $\mathbf{c}_i(t)$ and $\mathbf{h}_i(t)$, they follow distinct continuous dynamics due to different encoded information.  
Specifically, there is no constraint on the internal evolution of $\mathbf{c}_i(t)$, hence we model  $\frac{d\mathbf{c}_i(t)}{dt}$ by an MLP. 
Differently, due to the temporal decaying effect of  historical activities, we add constraints to the form of $\mathbf{h}_i(t)$ to model such an effect. Concretely, we use another MLP followed by a Softplus activation layer to model the decay rate. The modeling of derivatives can be formulated as follows:

\begin{align}
     &\frac{d\mathbf{c}_i(t)}{dt} = \text{MLP}(\mathbf{c}_i(t)\oplus\mathbf{h}_i(t))\ ,\\
     &\alpha_i = \sigma(\text{MLP}(\mathbf{c}_i(t))\ , \\
     &\frac{d\mathbf{h}_i(t)}{dt} = -\alpha_i\mathbf{h}_i(t)\ ,
\end{align}

\noindent where $\sigma$ is the Softplus activation function to guarantee a positive decay rate, and $\oplus$ denotes the vector concatenation.  

\textbf{\textit{Instantaneous jump.}}
The bottom part in Figure~\ref{fig:dynamic_net} illustrates the network design to model the instantaneous jump introduced by happened activities.
Specifically, the function $\omega_i$ in Eq.~\eqref{eq:derivative} outputs the effects of the instantaneous jump, and it is modeled by an MLP in practice. 
As discussed before, the vector $\mathbf{h}_i(t)$  encodes the activity memory, and thus it is reasonable that the instantaneous jump will only affect the vector $\mathbf{h}_i(t)$. As a result, an activity of the $i_{th}$ need level gives rise to a change $\Delta \mathbf{h}_i(t)$ only to the corresponding activity memory embedding $\mathbf{h}_i(t)$, \emph{i.e.}, $\Delta \mathbf{h}_j(t)=0, \forall j\neq i$ and $\Delta \mathbf{c}_i(t) = 0, \forall i$. The MLP takes in the concatenation of the activity embedding $\mathbf{k}(t)$ and the internal state $\mathbf{c}_i(t)$, and outputs the variation $\Delta \mathbf{h}_i(t)$ in the memory embedding $\mathbf{h}_i(t)$,  which is formulated as follows:

\begin{align}
    &\Delta \mathbf{h}_i(t) = \text{MLP}(\mathbf{k}(t) \oplus \mathbf{c}_i(t)) \ , \\
    &\lim\limits_{\epsilon\to 0^+}\mathbf{h}_i(t+\epsilon) = \mathbf{h}_i(t) + \Delta \mathbf{h}_i(t) \ ,
\end{align}

\noindent where $\mathbf{k}(t)$ denotes the activity associated with the $i_{\text{th}}$ need level.

\subsubsection{\textbf{Policy Function.}}
\begin{figure}[t]
    \centering
    \includegraphics[width=0.99\linewidth]{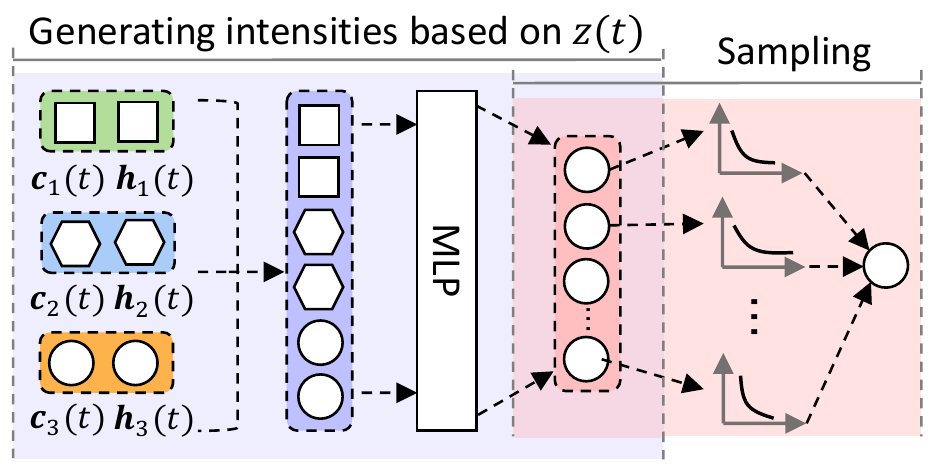}
    \caption{Network architecture of the policy function.}
    \label{fig:intensity}
\end{figure}

Based on the activity intensity function $\lambda_k(t)$, the probability of activity type $k$ happens within the time interval $[t, t+dt)$  is as:  $P\{\text{activity $k$ happens in} [t, t+dt)\} = \lambda_k(t) \cdot dt$.
The policy function is a mapping from the state to action that generates the arrival of the next activity with the type conditioned on the current state. With the modeling of activity intensities, the goal of the policy function is to generate intensities based on the need states $\mathbf{z}(t)$. Figure~\ref{fig:intensity} shows the network design of the policy function.  Although the three need levels control specific activities, they are not independent and can be pursued simultaneously, which may give rise to competing activity choices. Therefore, the states of the three levels all affect the generation of the next activity. In other words, the activity intensity $\lambda^*_k(t)$ is conditioned on embedding processes of all need levels. To model the interactions between different levels in determining the next activity, we concatenate the three embedding processes $\mathbf{z}_i(t), i\in\{1, 2, 3\}$ and leverage an MLP to obtain conditional activity intensities. Here we perform the sampling to obtain the time interval and the activity type based on the total condition intensity and type distribution as:

\begin{equation}
    \lambda^*(t) =\sum_{k=1}^M\lambda_k(t), \ \ 
    p(k|t) = \frac{\lambda_k(t)}{\sum_{k=1}^M\lambda_k(t)}
\end{equation}

\noindent where $M$ is the number of activity types.

\subsubsection{\textbf{Reward Function.}}
GAIL uses a reward function to evaluate the actions by comparing the generated state-action pairs with the real pairs, which is modeled by a discriminator network $D_\phi$. To compare the real and policy-generated pairs more effectively, we also utilize the historical sequence information, thus, the state in the discriminator is defined as $s_d = (\mathbf{z}(t),\mathbf{S})$. For the sequence $\mathbf{S}=[x_1,x_2,...,x_n]$, $x_i$ contains the information of the time interval $\tau_i$, hour $h_i$, weekday $w_i$, activity type $k_i$, and need level $n_i$. In addition, the action $a$ is set as the time interval $\tau$ since the last activity to the current one, \emph{i.e.}, $a=(\tau,k)$. Based on the above notations, the output of the discriminator can be defined as $D_\phi(s_d, a)$.  

Through an embedding layer, we first transform $s_d$ and $a$ into embeddings. Then we leverage an attention mechanism to aggregate the sequential features. The concatenation of the sequential embedding, state $\mathbf{z}(t)$, and action embedding is fed into an MLP with a sigmoid activation function. Thus, the reward function can be expressed as:  $R(s,a) = \text{log}D_\phi(s_d,a)$.

The training process including (i) GAIL training and (ii) pre-training is introduced in Appendix. The simulation algorithm is presented in Appendix.

\section{Experiments}

In this section, we conduct extensive experiments to investigate the following research problems:

\begin{itemize}[leftmargin=*]
  \item  \textbf{RQ1}: How does SAND perform in retaining the data fidelity and reflecting activity characteristics compared with baseline solutions?
 \item \textbf{RQ2}: How do different components of SAND contribute to the final performance?
 \item \textbf{RQ3}: Can SAND generate high-quality synthetic data that benefit practical applications?
 \item \textbf{RQ4}: Can SAND provide insightful interpretations on modeling daily activities?
\end{itemize}

\begin{table*}[t]
\centering
\caption{Overall performance of SAND and baselines in terms of the JSD-based metrics, and lower results are better. Bold denotes the best results and \underline{underline} denotes the second-best results. The improvements are significant (p-value<0.05).}\label{tbl:JSD}
\vspace{-3mm}
\resizebox{\textwidth}{!}{%
\begin{tabular}{|c|cccccc|cccccc|}
\hline
\textbf{Dataset}  & \multicolumn{6}{c|}{\textbf{Mobile Operator}} & \multicolumn{6}{c|}{\textbf{Foursquare}} \\
\hline
Metrics (JSD) & MacroInt & MicroInt & DailyAct & ActType &  Weekday & Hour & MacroInt & MicroInt & DailyAct & ActType & Weekday & Hour\\ 
\hline
Semi-Markov   & 0.291 & 0.158 & 0.439 & 0.471  & 0.0042 & 0.051 & 0.334 & 0.055 & 0.485 & 0.101 & 0.0032 & 0.051 \\
Hawkes & 0.276 & 0.151 & 0.542 & 0.123 & 0.0039 & 0.051 & 0.073 & \underline{0.024} & 0.530 & 0.026 & 0.0024 & 0.047\\
\hline
Neural Hawkes & 0.026 & 0.143 & 0.125 & \textbf{0.0063} & 0.0036 & 0.052 & 0.072 & 0.041 & 0.119 & 0.012 & 0.0040 & 0.047\\
Neural JSDE    & \underline{0.014}  & \underline{0.106} & 0.138 & 0.048 & \underline{0.0033} & 0.051 & \underline{0.041} & 0.033 &  \textbf{0.056} &  \underline{0.0072} & \underline{0.0022} & \underline{0.046} \\
THP  & 0.167 & 0.111 & 0.058 &  0.098 & 0.005 & 0.040 & 0.331 &0.035 & 0.095 & 0.003 &  0.013 & 0.047 \\
\hline
LSTM          & 0.110 & 0.136 & 0.513 & 0.342  & 0.0041 & 0.050 & 0.249 & 0.217 & 0.628 & 0.073 & 0.0033 & 0.051\\
SeqGAN          & 0.143 & 0.128 & 0.047 & 0.054  & 0.022  & 0.072  & 0.225 & 0.178 & 0.627 & 0.065 & 0.0034 & 0.051\\
GAIL  & 0.089 & 0.120 & \underline{0.040} & 0.231  & 0.005  & \underline{0.050}  & 0.226 & 0.118 & 0.167 & 0.087 & 0.0049 & 0.062\\
\hline
SAND          & \textbf{0.0096} & \textbf{0.084} & \textbf{0.025} & \underline{0.036} & \textbf{0.002} & \textbf{0.009}  & \textbf{0.018} & \textbf{0.014} & \underline{0.062} & \textbf{0.0044} & \textbf{0.00032} & \textbf{0.0069}\\ 
\hline
\end{tabular}%
}
\end{table*}

\subsection{Experimental Settings}

\subsubsection{\textbf{Datasets}}. We conduct extensive experiments  on two real-world datasets.  (1) Foursquare-NYC~\citep{yang2014modeling} dataset contains checkin activities to various POIs collected from 2000 users with 14 activity labels during the duration from 2012-05-01 to 2012-06-01. (2) Mobile dataset contains 10000 users with 15 activity labels during the duration from 2016-09-17 to 2016-10-17, which is collected in Beijing by a major mobile operator in China. 
We take careful steps to consider ethical issues in using data\footnote{First, the Terms of Service for both datasets include consent for research studies.  Second, the research protocol has been reviewed and approved by our local institutional board.  All research data is sanitized for privacy preservation, with limited access to authorized researchers bound by non-disclosure agreements.}.

\subsubsection{Need Annotation}\label{sec:annotation}
According to the definition and description of each need level in Section~\ref{sub:need_aware}, we ask three annotators to label each activity with one of the need levels. 
To ensure that correct expert knowledge is utilized, the three annotators all have expertise in related knowledge, including a senior Ph.D. candidate and two postdocs with a background in psychology and behavioral sciences. We choose the number of experts (three) following studies in NLP with annotation tasks~\citep{stab2014annotating}. If the three experts disagree on the label, we will invite another expert and start a discussion. Through this process, all activities obtain consistent labels. The annotation approach has satisfied the requirement of our problem settings due to the small scale of activity types.

\subsubsection{\textbf{Baselines.}} To evaluate the performance of the SAND framework, we compare it against state-of-the-art baseline methods: 
Semi-Markov~\cite{limnios2012semi}, a classical probability model;
Hawkes Process~\cite{laub2015hawkes}, a representative point process model;
Neural Hawkes Process~\cite{mei2016neural}, the neural extension to the Hawkes process;
Transformer Hawkes Process~\citep{zuo2020transformer} (THP) is another neural extension to the Hawkes process, which utilizes the self-attention mechanism to capture long-term dependencies;
Neural JSDE~\cite{chen2018neural}, the state-of-the-art method to learn continuous and discrete dynamic behavior;
LSTM~\cite{hochreiter1997long}, a widely used model in sequence prediction;
SeqGAN~\cite{yu2017seqgan}, the state-of-the-art model for discrete sequence generation;
TrajGAIL~\cite{ho2016generative}, a model-free imitation learning algorithm in trajectory generation.

\subsubsection{\textbf{Metrics.}} 
We measure whether synthetic data accurately reflects crucial characteristics of the original, real-world data.
Following the mainstream practice in previous works~\citep{feng2020learning,ouyang2018non}, we use essential metrics to describe activity patterns for comparing the statistical similarity between the generated data and real-world data, including (1) \textit{ActInt}: time intervals between activities, including type-free intervals~(MacroInt) and type-aware intervals~(MicroInt); (2) \textit{DailyAct}: daily happened activities. It is the number of activities in one day for each individual;  (3) \textit{ActType}: the overall distribution over different activity types; (4) \textit{Weekday}: the overall time distribution over the seven days; (5) \textit{Hour}: the overall time distribution  over the twenty-four hours.
To get the quantitative evaluations on the fidelity of generated data, we use Jensen–Shannon divergence~($JSD$) to measure the distribution similarity of the above patterns between the generated data and real-world data, which is a widely used distance metric for comparing two distributions as follows:
\begin{equation}
    \text{JSD}(P||Q) = H(M)-\frac{1}{2}(H(P)+H(Q))
\end{equation}
\noindent where $H$ is the Shannon entropy, $p$ and $q$ are distributions, and $M=\frac{p+q}{2}$. In our setup, lower JSD denotes a closer distribution between synthetic data and real data, which indicates a better generative model. In addition, the JSD is bounded by $[0,1]$ for two probability distributions with the base 2 logarithm~\citep{lin1991divergence}.

\subsection{Overall Performance (RQ1)}\label{sec:overall_performance}
 Table~\ref{tbl:JSD} reports the performance in retaining the data fidelity of our framework and the eight competitive baselines on two real-world datasets. 
From the results, we have the following findings:

\begin{table*}[t]
\centering
\caption{Ablation study on SAND variants. Bold denotes the best results and \underline{underline} denotes the second-best results.}\label{tbl:ablation}
\vspace{-3mm}
\resizebox{\textwidth}{!}{%
\begin{tabular}{|c|cccccc|cccccc|}
\hline
\textbf{Dataset}  & \multicolumn{6}{c|}{\textbf{Mobile Operator}} & \multicolumn{6}{c|}{\textbf{Foursquare}} \\
\hline
Metrics (JSD) & MacroInt & MicroInt & DailyAct & ActType &  Weekday & Hour & MacroInt & MicroInt & DailyAct & ActType & Weekday & Hour\\ 
\hline
SAND          & \textbf{0.013} & \textbf{0.084} & \textbf{0.025} & \textbf{0.036} & \textbf{0.002} & \textbf{0.009}  & \textbf{0.018} & \textbf{0.014} & \underline{0.062} & \textbf{0.0044} & \textbf{0.00032} & \textbf{0.0069}\\ 
\hline
SAND - GAIL & \underline{0.013} & 0.116 & 0.085 & 0.040 & \underline{0.0031} & 0.051 & 0.039 & \underline{0.028} & 0.202 & \underline{0.0051} & \underline{0.0018} &  \underline{0.0092}  \\
\hline
SAND - need & 0.014 & 0.116 & 0.085 & \underline{0.039} & 0.0035 & \underline{0.050} & \underline{0.019} & 0.030 & \textbf{0.0085} & 0.0072 & 0.0021 & 0.048\\
\hline
SAND - pretrain & 0.015 & \underline{0.110} & \underline{0.059} & 0.190 & 0.004 & \underline{0.048} & 0.070 & 0.025 & 0.161 & 0.064 & 0.0020 & 0.044 \\
\hline
\end{tabular}%
}
\end{table*}

\begin{itemize}[itemsep=2pt,topsep=0pt,parsep=0pt,leftmargin=*]
    \item \textbf{Our framework steadily achieves the best performance.} SAND achieves the best performance on the mobile operator dataset, by ranking first on five metrics and second on one metric.  For five metrics that rank 1st, SAND reduces the JSD by more than 20\%. It also shows superior performance on most of the metrics on the Foursquare dataset, which ranks first on five metrics by reducing JSD by more than 40\%. Meanwhile, it achieves comparable performance with the best baseline on the other one metric.
    \item \textbf{Time-invariant model performs poorly in simulating human activities.}  Semi-Markov performs the worst in most cases. which indicates that the time-invariant assumption fails to describe behavior transition laws due to the existence of  complex temporal patterns in daily activities.
    \item \textbf{Learning from raw data alone is insufficient for a realistic simulation.} The LSTM model has a poor performance on the metrics of \textit{DailyAct} and \textit{ActType}, which means errors can be accumulated in the step-by-step generation process.  By contrast, SeqGAN and GAIL improve the performance by using reinforcement learning and adversarial learning. For the Foursquare dataset that is more sparse, their superiority is lost, which further suggests the instability of purely data-driven methods.
    \item \textbf{It is essential to model dynamic human needs.} The neural Hawkes, THP, and neural JSDE almost achieve the sub-optimal results on the two datasets, indicating the rationality of characterizing events in continuous time by temporal point processes.  However, without investigating the deeper mechanism behind observed activities, their performance is still limited.
\end{itemize}

\begin{figure}[t]
    \centering
    \subfigure[Mobile Operator Dataset.]{
        \label{fig:prediction_bj}
        \includegraphics[width=0.486\columnwidth]{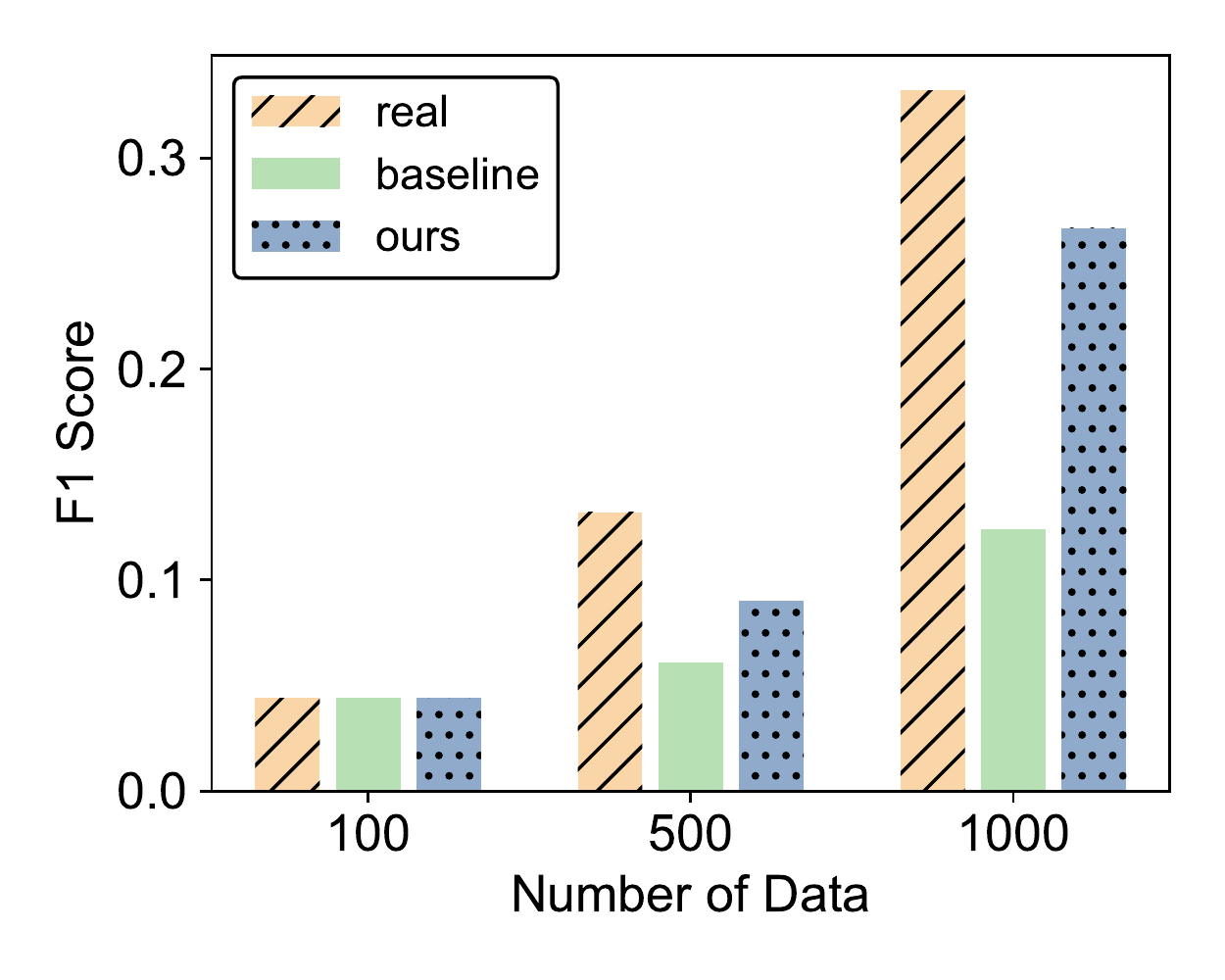}}
    \subfigure[Foursquare Dataset.]{
        \label{fig:prediction_usa}
        \includegraphics[width=0.486\columnwidth]{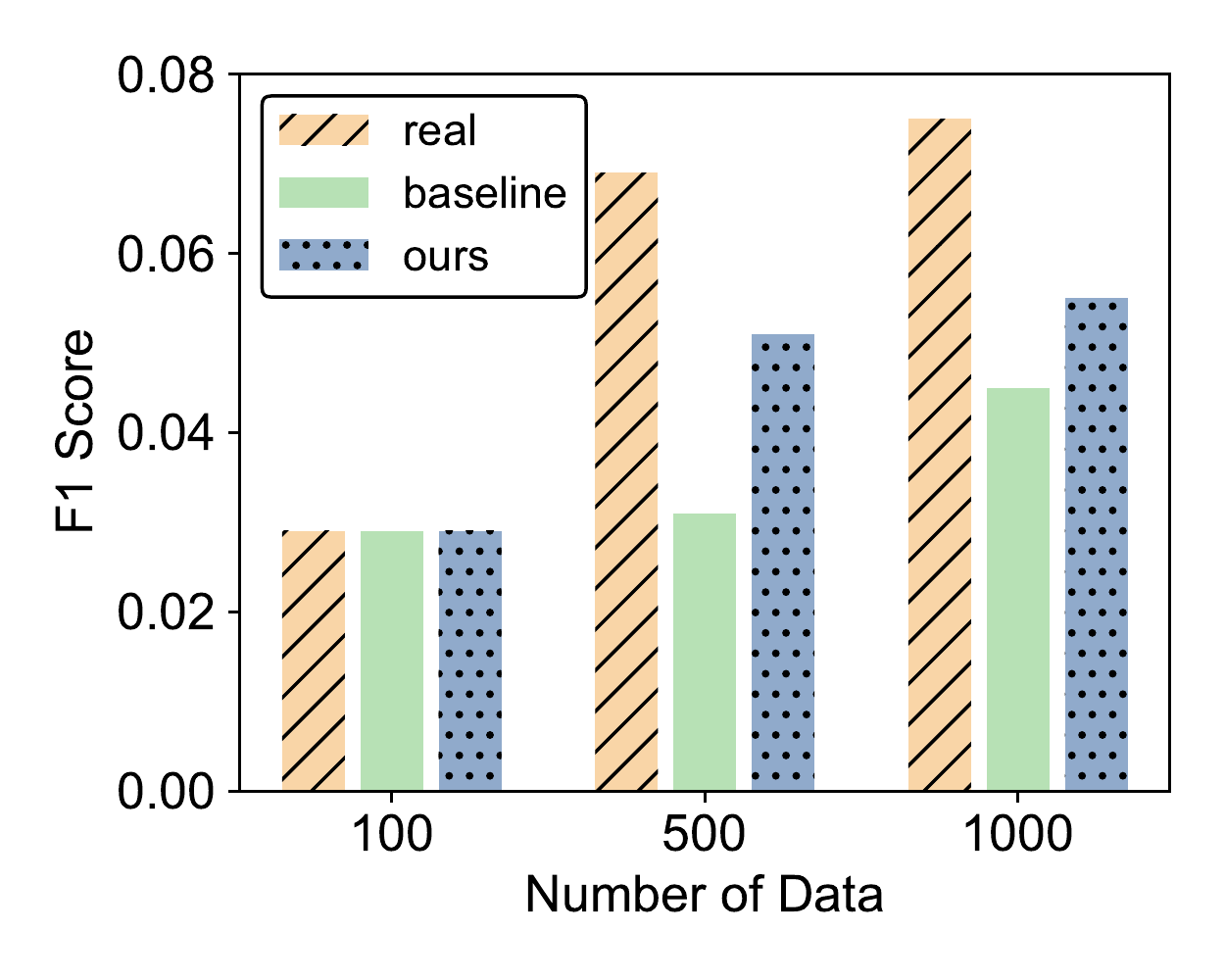}}
    \vspace{-2mm}
    \caption{Activity prediction in the fully synthetic scenario.}
    \label{fig:prediction_fully}
    \vspace{-3mm}
\end{figure}

\begin{figure}[t]
    \centering
    \subfigure[Mobile Operator Dataset.]{
        \label{fig:prediction_bj}
        \includegraphics[width=0.486\columnwidth]{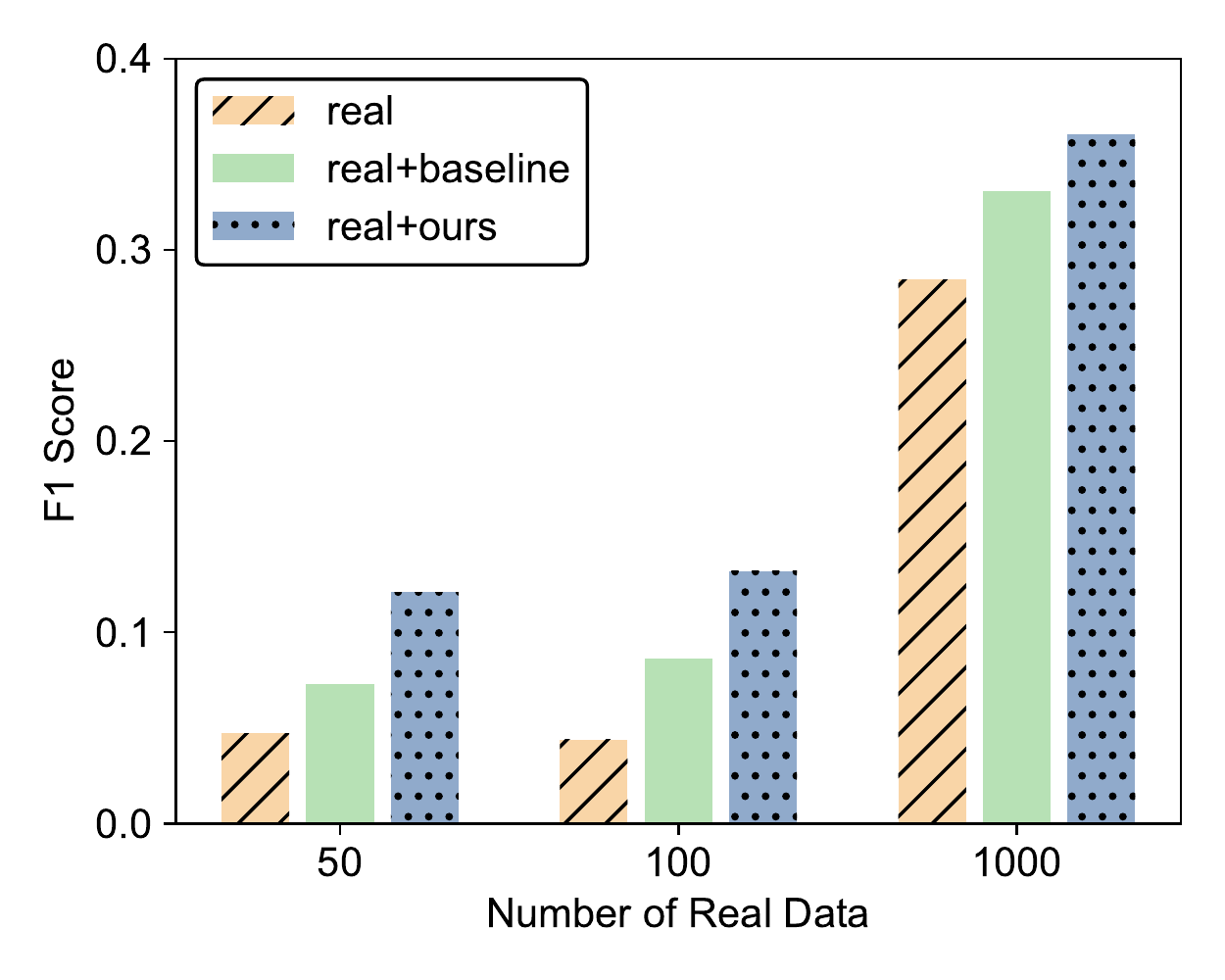}}
    \subfigure[Foursquare Dataset.]{
        \label{fig:prediction_usa}
        \includegraphics[width=0.486\columnwidth]{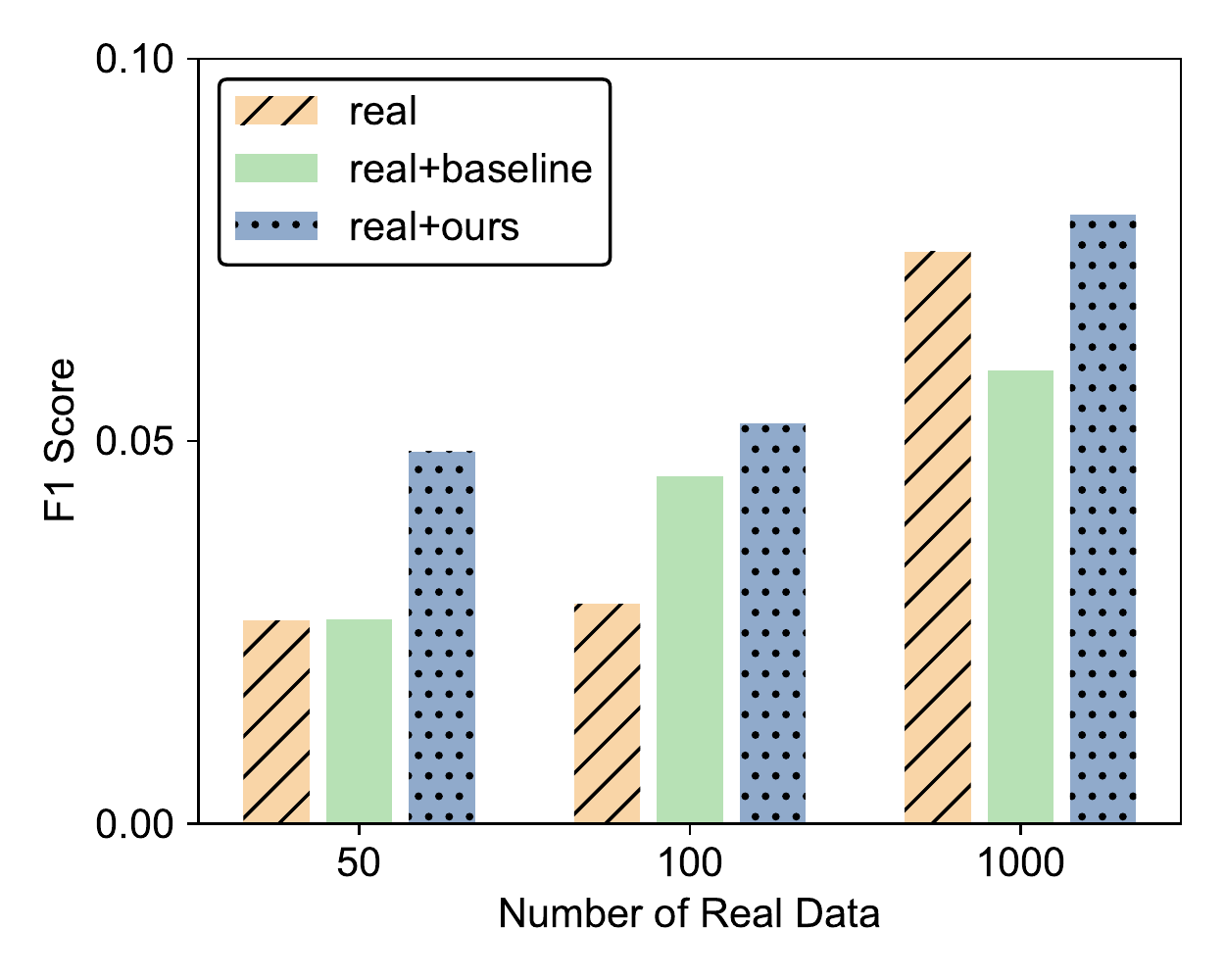}}
    \vspace{-2mm}
    \caption{Activity prediction in the hybrid scenario. For a different number of real-world sequences, \emph{i.e.}, 50, 100, 1000, we all add 1000 generated sequences for data augmentation.}
    \label{fig:prediction}
    \vspace{-3mm}
\end{figure}

\subsection{Ablation Studies (RQ2)}

The proposed SAND framework consists of two key components: modeling need dynamics and solving the MDPs with GAIL.  Besides, we also use the pre-training mechanism.
To further validate whether they are indeed crucial for the final performance,  we conduct ablation studies on two datasets by comparing the performance of three variants of SAND, including \textit{SAND - need, SAND - GAIL, SAND - pretrain}. Specifically, \textit{SAND - need}  calculates the latent state as~\citep{jia2019neural} without modeling hierarchical human needs,  \textit{SAND - GAIL} removes the GAIL training framework, and \textit{SAND - pretrain} starts training from raw data without the pre-training mechanism. 

The evaluation results are reported in Table~\ref{tbl:ablation}. We can observe that SAND delivers the best performance on five metrics compared with the variants that are removed with specific designs.  Without modeling need dynamics, the performance is reduced significantly, indicating the necessity to consider the intrinsic motivation in human activity simulation.  Besides, removing the GAIL framework also reduces the data fidelity, which suggests the strong modeling capabilities of generative adversarial mechanisms.  In addition, the pre-training mechanism facilitates making full use of the activity data and enables our framework to preview the dependencies and regularities of daily activities before GAIL training, thus it also contributes to the final performance.

\subsection{Practical Applications (RQ3)}\label{sec:apply}

In user-based applications, real-world activity records usually cannot be directly shared due to privacy issues. Under this circumstance, SAND can be used to generate synthetic data to mask sensitive information while retaining the usability of real data. To examine the utility of the generated synthetic data, we perform experiments with synthetic data of two categories:
\begin{itemize}[leftmargin=*]
    \item \textbf{Fully synthetic scenario}; Only synthetic data is used in applications, which provides a more robust privacy protection.
    \item \textbf{Hybrid scenario}; It combines real and synthetic data, which is widely used in data augmentation settings.
\end{itemize}
We select two representative applications~\citep{jakkula2007mining,minor2015data} based on the activity data: (1) activity prediction and (2) interval estimation, which are fundamental to many activity-related problems, such as activity recommendation and planning.

\begin{figure*}[t!]
    \centering
    \hspace{-2mm}
    \subfigure[Individual 1]{
        \label{fig:case1}
        \includegraphics[width=0.92\columnwidth]{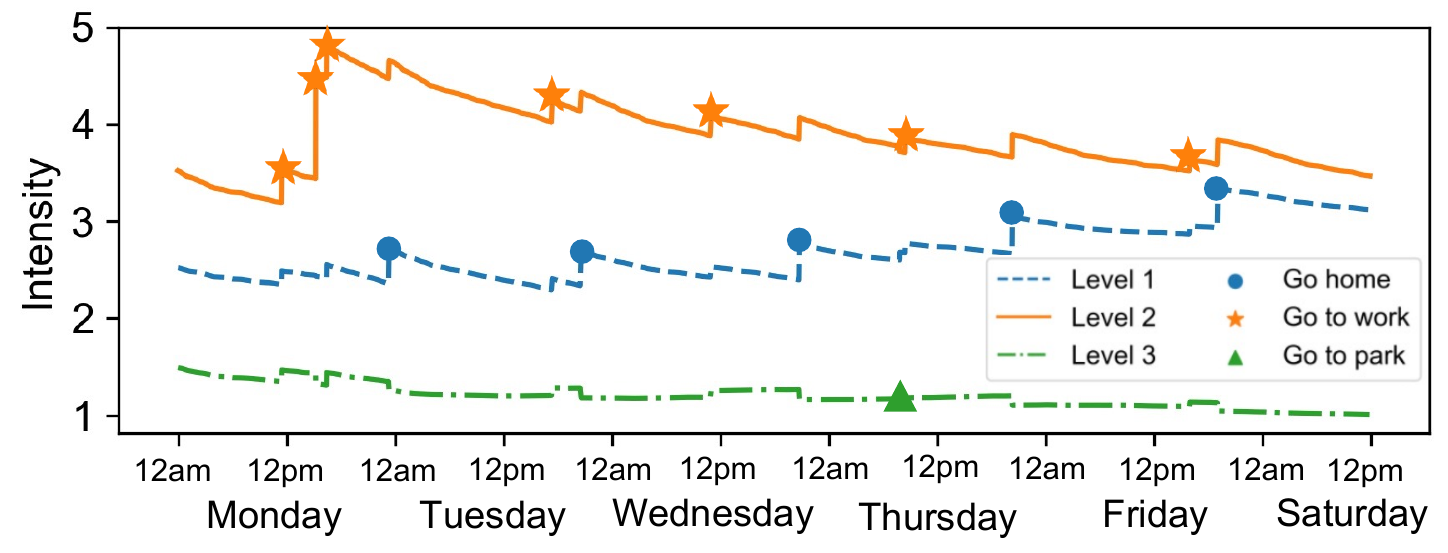}}
    \hspace{+5mm}
    \subfigure[Individual 2]{
        \label{fig:case2}
        \includegraphics[width=0.93\columnwidth]{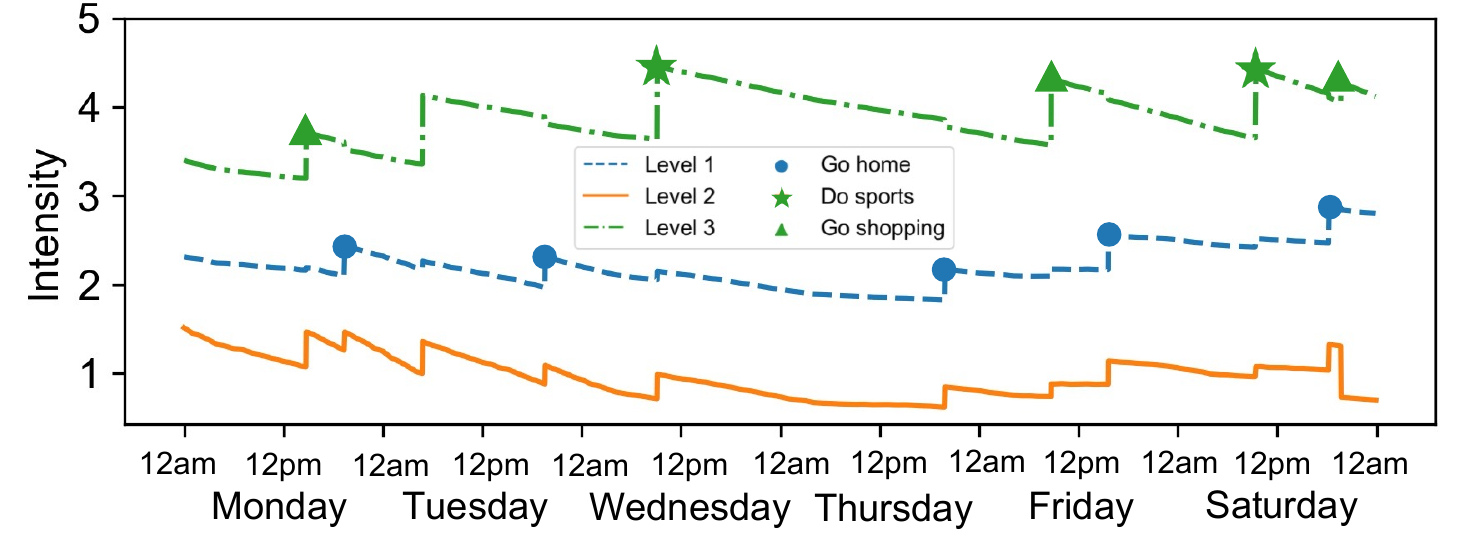}}
    \vspace{-3mm}
    \caption{Case study of two generated activity sequences and the learned intensity of different need levels. We select two representative individuals with different activity patterns.}
    \label{fig:cases}
    \vspace{-3mm}
\end{figure*}

We utilize a widely-used model, LSTM with attention mechanism, to predict individuals' future activity types based on their historical sequence. 
As shown in Figure~\ref{fig:prediction_fully}, compared with the best baseline, the prediction performance on the dataset generated by our framework is much closer to the performance on the real data, showing the retained utility of the generated data.
Figure~\ref{fig:prediction} illustrates that the model trained on the augmented data exhibits significantly better performance than that only trained on the real-world data.  Meanwhile, the data augmented by SAND outperforms that by the best baseline. Moreover, the augmented data becomes more useful when the real-world data is of small scale, \emph{e.g.}, only with 50 or 100 real-world sequences.  These results validate the practical value of the synthetic data.

\subsection{Interpretability of Dynamic Needs (RQ4)}\label{sec:interpret}

To validate whether SAND can provide insightful interpretability, we perform a case study on the learned intensity values of different need levels in the simulation process.  Figure~\ref{fig:cases} illustrates the simulated activity sequences of two individuals for one week, together with the corresponding intensity values of three need levels. 
In terms of the model interpretability, we have two main observations. First, the proposed SAND can generate distinct but lifelike activity sequences that are hard to tell apart from real-world data.
Specifically, comparing Figure~\ref{fig:cases}(a) and (b), the two synthetic individuals lead quite personalized lifestyles. Individual 1 follows regular working routines with the intensity dynamics of the level-2 need varying periodically, while individual 2 enjoys more freedom without working, showing a constantly low intensity of the level-2 need. 
Second, SAND can simulate human daily activity in an interpretable way with need modeling.
As observed from Figure~\ref{fig:cases}, the occurrence of activity not only changes the intensity of the corresponding need level but also affects other levels, indicating
that different need levels are interconnected by intensities derived from need states and trigger activities in a cooperative manner.
In summary, the above observations demonstrate the interpretability of SAND for simulation outcomes, which is equally important in real-life applications.

\section{Related Work}

\noindent\textbf{Human activity simulation.}
Solutions for activity simulation are mainly agent-based modeling~\cite{macal2005tutorial} with rule-based methods~\cite{kavak2019location,kim2020location,kim2019simulating,pfosertowards,muric2020massive,north2010multiscale}. Specifically, these methods assume that human activities can be described by limited parameters with explicit physical meaning and are governed by transition rules based on psychology and social science theories. With simplified assumptions of human behaviors, agents in the system can be assigned different goals, then they take actions to maximize different attributes. For example, Kim et al.~\cite{kim2019simulating} propose that human actions are triggered by a cause and give rise to corresponding effects. Besides, considering the multiple behaviors, the priorities of behaviors are determined based on Maslow's hierarchy of needs~\cite{kim2019simulating,kavak2019location,kim2020location}. 
Despite the promising performance under some circumstances, rule-based methods fail to capture complicated activity patterns due to relying on simplified assumptions and thus usually fail to simulate activities in reality.  The purpose of activity simulation is different from that of activity prediction~\cite{minor2015data,jakkula2007mining,ye2013s}. The former emphasizes the simulation results to reproduce and reflect characteristics of real data, but should not be too similar to real data with the goal of protecting user privacy, while the latter highlights to what extent the model can recover the real data. Although deep learning approaches are proposed for activity prediction~\cite{nweke2018deep,jaouedi2020prediction}, the problem of simulating daily activities has been barely explored.

\noindent\textbf{Deep generative models for activity simulation.}
Deep generative models, such as generative adversarial networks (GAN)~\cite{goodfellow2014generative} and variational autoencoder (VAE)~\cite{kingma2019introduction},  are promising solutions to simulation. Previous studies~\cite{wei2021we,pan2020xgail,zhang2019unveiling,shi2019virtual} have also explored the ability of Generative adversarial Imitation Learning (GAIL) to simulate human decision process.  Besides, a series of neural temporal point process models~\cite{mei2016neural,zuo2020transformer,chen2018neural,yuan2022activity} are proposed to model discrete events. Although these models are mainly for discrete event prediction, the learned probability distribution provides opportunities to perform event generation by the sampling operation.  Recently, Gupta et al.~\cite{gupta2022proactive} propose attention-based temporal point process flows to model goal-directed activity sequences. However, it is not appropriate for our research problems as daily activities cannot be represented as a sequence of actions performed to achieve an explicit goal. We propose a knowledge-driven framework based on GAIL, and the incorporation of psychological knowledge is realized by leveraging an ODE-based temporal point process.

\section{conclusion}
In this paper, we investigate the individual activity simulation problem by proposing a novel framework SAND, which integrates deep generative models with well-respected psychological theories. 
Extensive experiments on two real-world datasets show the superior performance of the proposed framework. 
Our framework is not strictly limited to Maslow's theories, instead, what we highlight is leveraging neural networks to learn the driving force behind human daily activities, and the choice of knowledge or theory related to such driving force is quite flexible. 
Importantly, effective modeling of human needs makes it possible to understand human behaviors at a deeper level, which not only benefits the activity simulation in this work but also contributes to many other problems of psychology-informed user modeling. 
In terms of limitations,  we recognize that data-driven models largely depend on high-quality datasets. For example, the shortage of long-term and fine-grained datasets hinders the modeling of needs for esteem and self-actualization.

\begin{acks}
This work was supported in part by the National Key Research and Development Program of China under grant 2020YFA0711403, the National Nature Science Foundation of China under 61971267, 61972223,  62171260, and U1936217, the Young Elite Scientists Sponsorship Program by CIC under 2021QNRC001,the Guoqiang Institute, Tsinghua University under 2021GQG1005, and Beijing National Research Center for Information Science and Technology (BNRist).
\end{acks}

\clearpage

\bibliographystyle{ACM-Reference-Format}
\bibliography{reference}


\begin{thebibliography}{52}


\ifx \showCODEN    \undefined \def \showCODEN     #1{\unskip}     \fi
\ifx \showDOI      \undefined \def \showDOI       #1{#1}\fi
\ifx \showISBNx    \undefined \def \showISBNx     #1{\unskip}     \fi
\ifx \showISBNxiii \undefined \def \showISBNxiii  #1{\unskip}     \fi
\ifx \showISSN     \undefined \def \showISSN      #1{\unskip}     \fi
\ifx \showLCCN     \undefined \def \showLCCN      #1{\unskip}     \fi
\ifx \shownote     \undefined \def \shownote      #1{#1}          \fi
\ifx \showarticletitle \undefined \def \showarticletitle #1{#1}   \fi
\ifx \showURL      \undefined \def \showURL       {\relax}        \fi
\providecommand\bibfield[2]{#2}
\providecommand\bibinfo[2]{#2}
\providecommand\natexlab[1]{#1}
\providecommand\showeprint[2][]{arXiv:#2}

\bibitem[\protect\citeauthoryear{Arentze, Hofman, van Mourik, and
  Timmermans}{Arentze et~al\mbox{.}}{2000}]%
        {arentze2000albatross}
\bibfield{author}{\bibinfo{person}{Theo Arentze}, \bibinfo{person}{Frank
  Hofman}, \bibinfo{person}{Henk van Mourik}, {and} \bibinfo{person}{Harry
  Timmermans}.} \bibinfo{year}{2000}\natexlab{}.
\newblock \showarticletitle{ALBATROSS: multiagent, rule-based model of activity
  pattern decisions}.
\newblock \bibinfo{journal}{\emph{Transportation Research Record}}
  \bibinfo{volume}{1706}, \bibinfo{number}{1} (\bibinfo{year}{2000}),
  \bibinfo{pages}{136--144}.
\newblock


\bibitem[\protect\citeauthoryear{Auld and Mohammadian}{Auld and
  Mohammadian}{2012}]%
        {auld2012activity}
\bibfield{author}{\bibinfo{person}{Joshua Auld} {and}
  \bibinfo{person}{Abolfazl~Kouros Mohammadian}.}
  \bibinfo{year}{2012}\natexlab{}.
\newblock \showarticletitle{Activity planning processes in the Agent-based
  Dynamic Activity Planning and Travel Scheduling (ADAPTS) model}.
\newblock \bibinfo{journal}{\emph{Transp Res Part A Policy Pract}}
  \bibinfo{volume}{46}, \bibinfo{number}{8} (\bibinfo{year}{2012}),
  \bibinfo{pages}{1386--1403}.
\newblock


\bibitem[\protect\citeauthoryear{Bowman and Ben-Akiva}{Bowman and
  Ben-Akiva}{2001}]%
        {bowman2001activity}
\bibfield{author}{\bibinfo{person}{John~L Bowman} {and}
  \bibinfo{person}{Moshe~E Ben-Akiva}.} \bibinfo{year}{2001}\natexlab{}.
\newblock \showarticletitle{Activity-based disaggregate travel demand model
  system with activity schedules}.
\newblock \bibinfo{journal}{\emph{Transportation research part a: policy and
  practice}} \bibinfo{volume}{35}, \bibinfo{number}{1} (\bibinfo{year}{2001}),
  \bibinfo{pages}{1--28}.
\newblock


\bibitem[\protect\citeauthoryear{Chang and Yuan}{Chang and Yuan}{2008}]%
        {chang2008synthesized}
\bibfield{author}{\bibinfo{person}{Wei-Lun Chang} {and}
  \bibinfo{person}{Soe-Tsyr Yuan}.} \bibinfo{year}{2008}\natexlab{}.
\newblock \showarticletitle{A synthesized model of Markov chain and ERG theory
  for behavior forecast in collaborative prototyping}.
\newblock \bibinfo{journal}{\emph{JITTA}} \bibinfo{volume}{9},
  \bibinfo{number}{2} (\bibinfo{year}{2008}), \bibinfo{pages}{5}.
\newblock


\bibitem[\protect\citeauthoryear{Chen, Rubanova, Bettencourt, and
  Duvenaud}{Chen et~al\mbox{.}}{2018}]%
        {chen2018neural}
\bibfield{author}{\bibinfo{person}{Ricky~TQ Chen}, \bibinfo{person}{Yulia
  Rubanova}, \bibinfo{person}{Jesse Bettencourt}, {and} \bibinfo{person}{David
  Duvenaud}.} \bibinfo{year}{2018}\natexlab{}.
\newblock \showarticletitle{Neural ordinary differential equations}.
\newblock \bibinfo{journal}{\emph{arXiv preprint arXiv:1806.07366}}
  (\bibinfo{year}{2018}).
\newblock


\bibitem[\protect\citeauthoryear{Chung}{Chung}{1969}]%
        {chung1969markov}
\bibfield{author}{\bibinfo{person}{Kae~H Chung}.}
  \bibinfo{year}{1969}\natexlab{}.
\newblock \showarticletitle{A Markov chain model of human needs: An extension
  of Maslow's need theory}.
\newblock \bibinfo{journal}{\emph{Academy of Management Journal}}
  \bibinfo{volume}{12}, \bibinfo{number}{2} (\bibinfo{year}{1969}),
  \bibinfo{pages}{223--234}.
\newblock


\bibitem[\protect\citeauthoryear{Ettema, Borgers, and Timmermans}{Ettema
  et~al\mbox{.}}{1993}]%
        {ettema1993simulation}
\bibfield{author}{\bibinfo{person}{Dick Ettema}, \bibinfo{person}{Aloys
  Borgers}, {and} \bibinfo{person}{Harry Timmermans}.}
  \bibinfo{year}{1993}\natexlab{}.
\newblock \showarticletitle{Simulation model of activity scheduling behavior}.
\newblock \bibinfo{journal}{\emph{Transportation Research Record}}
  (\bibinfo{year}{1993}), \bibinfo{pages}{1--1}.
\newblock


\bibitem[\protect\citeauthoryear{Feng, Yang, Xu, Yu, Wang, and Li}{Feng
  et~al\mbox{.}}{2020}]%
        {feng2020learning}
\bibfield{author}{\bibinfo{person}{Jie Feng}, \bibinfo{person}{Zeyu Yang},
  \bibinfo{person}{Fengli Xu}, \bibinfo{person}{Haisu Yu},
  \bibinfo{person}{Mudan Wang}, {and} \bibinfo{person}{Yong Li}.}
  \bibinfo{year}{2020}\natexlab{}.
\newblock \showarticletitle{Learning to simulate human mobility}. In
  \bibinfo{booktitle}{\emph{KDD}}. \bibinfo{pages}{3426--3433}.
\newblock


\bibitem[\protect\citeauthoryear{Goodfellow, Pouget-Abadie, Mirza, Xu,
  Warde-Farley, Ozair, Courville, and Bengio}{Goodfellow et~al\mbox{.}}{2014}]%
        {goodfellow2014generative}
\bibfield{author}{\bibinfo{person}{Ian Goodfellow}, \bibinfo{person}{Jean
  Pouget-Abadie}, \bibinfo{person}{Mehdi Mirza}, \bibinfo{person}{Bing Xu},
  \bibinfo{person}{David Warde-Farley}, \bibinfo{person}{Sherjil Ozair},
  \bibinfo{person}{Aaron Courville}, {and} \bibinfo{person}{Yoshua Bengio}.}
  \bibinfo{year}{2014}\natexlab{}.
\newblock \showarticletitle{Generative adversarial nets}.
\newblock \bibinfo{journal}{\emph{NIPS}}  \bibinfo{volume}{27}
  (\bibinfo{year}{2014}).
\newblock


\bibitem[\protect\citeauthoryear{Graves}{Graves}{2013}]%
        {graves2013generating}
\bibfield{author}{\bibinfo{person}{Alex Graves}.}
  \bibinfo{year}{2013}\natexlab{}.
\newblock \showarticletitle{Generating sequences with recurrent neural
  networks}.
\newblock \bibinfo{journal}{\emph{arXiv preprint arXiv:1308.0850}}
  (\bibinfo{year}{2013}).
\newblock


\bibitem[\protect\citeauthoryear{Gupta and Bedathur}{Gupta and
  Bedathur}{2022}]%
        {gupta2022proactive}
\bibfield{author}{\bibinfo{person}{Vinayak Gupta} {and}
  \bibinfo{person}{Srikanta Bedathur}.} \bibinfo{year}{2022}\natexlab{}.
\newblock \showarticletitle{ProActive: Self-attentive temporal point process
  flows for activity sequences}. In \bibinfo{booktitle}{\emph{Proceedings of
  the 28th ACM SIGKDD Conference on Knowledge Discovery and Data Mining}}.
  \bibinfo{pages}{496--504}.
\newblock


\bibitem[\protect\citeauthoryear{Ho and Ermon}{Ho and Ermon}{2016}]%
        {ho2016generative}
\bibfield{author}{\bibinfo{person}{Jonathan Ho} {and} \bibinfo{person}{Stefano
  Ermon}.} \bibinfo{year}{2016}\natexlab{}.
\newblock \showarticletitle{Generative adversarial imitation learning}.
\newblock \bibinfo{journal}{\emph{Advances in neural information processing
  systems}}  \bibinfo{volume}{29} (\bibinfo{year}{2016}),
  \bibinfo{pages}{4565--4573}.
\newblock


\bibitem[\protect\citeauthoryear{Hochreiter and Schmidhuber}{Hochreiter and
  Schmidhuber}{1997}]%
        {hochreiter1997long}
\bibfield{author}{\bibinfo{person}{Sepp Hochreiter} {and}
  \bibinfo{person}{J{\"u}rgen Schmidhuber}.} \bibinfo{year}{1997}\natexlab{}.
\newblock \showarticletitle{Long short-term memory}.
\newblock \bibinfo{journal}{\emph{Neural computation}} \bibinfo{volume}{9},
  \bibinfo{number}{8} (\bibinfo{year}{1997}), \bibinfo{pages}{1735--1780}.
\newblock


\bibitem[\protect\citeauthoryear{Jakkula and Cook}{Jakkula and Cook}{2007}]%
        {jakkula2007mining}
\bibfield{author}{\bibinfo{person}{Vikramaditya Jakkula} {and}
  \bibinfo{person}{Diane~J Cook}.} \bibinfo{year}{2007}\natexlab{}.
\newblock \showarticletitle{Mining sensor data in smart environment for
  temporal activity prediction}.
\newblock \bibinfo{journal}{\emph{KDD}} (\bibinfo{year}{2007}).
\newblock


\bibitem[\protect\citeauthoryear{Jaouedi, Perales, Buades, Boujnah, and
  Bouhlel}{Jaouedi et~al\mbox{.}}{2020}]%
        {jaouedi2020prediction}
\bibfield{author}{\bibinfo{person}{Neziha Jaouedi},
  \bibinfo{person}{Francisco~J Perales}, \bibinfo{person}{Jos{\'e}~Maria
  Buades}, \bibinfo{person}{Noureddine Boujnah}, {and}
  \bibinfo{person}{Med~Salim Bouhlel}.} \bibinfo{year}{2020}\natexlab{}.
\newblock \showarticletitle{Prediction of human activities based on a new
  structure of skeleton features and deep learning model}.
\newblock \bibinfo{journal}{\emph{Sensors}} \bibinfo{volume}{20},
  \bibinfo{number}{17} (\bibinfo{year}{2020}), \bibinfo{pages}{4944}.
\newblock


\bibitem[\protect\citeauthoryear{Jia and Benson}{Jia and Benson}{2019}]%
        {jia2019neural}
\bibfield{author}{\bibinfo{person}{Junteng Jia} {and} \bibinfo{person}{Austin~R
  Benson}.} \bibinfo{year}{2019}\natexlab{}.
\newblock \showarticletitle{Neural jump stochastic differential equations}.
\newblock \bibinfo{journal}{\emph{arXiv preprint arXiv:1905.10403}}
  (\bibinfo{year}{2019}).
\newblock


\bibitem[\protect\citeauthoryear{Katz}{Katz}{1983}]%
        {katz1983assessing}
\bibfield{author}{\bibinfo{person}{Sidney Katz}.}
  \bibinfo{year}{1983}\natexlab{}.
\newblock \showarticletitle{Assessing self-maintenance: activities of daily
  living, mobility, and instrumental activities of daily living}.
\newblock \bibinfo{journal}{\emph{Journal of the American Geriatrics Society}}
  \bibinfo{volume}{31}, \bibinfo{number}{12} (\bibinfo{year}{1983}),
  \bibinfo{pages}{721--727}.
\newblock


\bibitem[\protect\citeauthoryear{Kavak, Kim, Crooks, Pfoser, Wenk, and
  Z{\"u}fle}{Kavak et~al\mbox{.}}{2019}]%
        {kavak2019location}
\bibfield{author}{\bibinfo{person}{Hamdi Kavak}, \bibinfo{person}{Joon-Seok
  Kim}, \bibinfo{person}{Andrew Crooks}, \bibinfo{person}{Dieter Pfoser},
  \bibinfo{person}{Carola Wenk}, {and} \bibinfo{person}{Andreas Z{\"u}fle}.}
  \bibinfo{year}{2019}\natexlab{}.
\newblock \showarticletitle{Location-based social simulation}. In
  \bibinfo{booktitle}{\emph{SSTD}}. \bibinfo{pages}{218--221}.
\newblock


\bibitem[\protect\citeauthoryear{Kim, Jin, Kavak, Rouly, Crooks, Pfoser, Wenk,
  and Z{\"u}fle}{Kim et~al\mbox{.}}{2020}]%
        {kim2020location}
\bibfield{author}{\bibinfo{person}{Joon-Seok Kim}, \bibinfo{person}{Hyunjee
  Jin}, \bibinfo{person}{Hamdi Kavak}, \bibinfo{person}{Ovi~Chris Rouly},
  \bibinfo{person}{Andrew Crooks}, \bibinfo{person}{Dieter Pfoser},
  \bibinfo{person}{Carola Wenk}, {and} \bibinfo{person}{Andreas Z{\"u}fle}.}
  \bibinfo{year}{2020}\natexlab{}.
\newblock \showarticletitle{Location-based social network data generation based
  on patterns of life}. In \bibinfo{booktitle}{\emph{MDM}}. IEEE,
  \bibinfo{pages}{158--167}.
\newblock


\bibitem[\protect\citeauthoryear{Kim, Kavak, Manzoor, Crooks, Pfoser, Wenk, and
  Z{\"u}fle}{Kim et~al\mbox{.}}{2019}]%
        {kim2019simulating}
\bibfield{author}{\bibinfo{person}{Joon-Seok Kim}, \bibinfo{person}{Hamdi
  Kavak}, \bibinfo{person}{Umar Manzoor}, \bibinfo{person}{Andrew Crooks},
  \bibinfo{person}{Dieter Pfoser}, \bibinfo{person}{Carola Wenk}, {and}
  \bibinfo{person}{Andreas Z{\"u}fle}.} \bibinfo{year}{2019}\natexlab{}.
\newblock \showarticletitle{Simulating urban patterns of life: A geo-social
  data generation framework}. In \bibinfo{booktitle}{\emph{SIGSPATIAL}}.
  \bibinfo{pages}{576--579}.
\newblock


\bibitem[\protect\citeauthoryear{Kingma, Welling, et~al\mbox{.}}{Kingma
  et~al\mbox{.}}{2019}]%
        {kingma2019introduction}
\bibfield{author}{\bibinfo{person}{Diederik~P Kingma}, \bibinfo{person}{Max
  Welling}, {et~al\mbox{.}}} \bibinfo{year}{2019}\natexlab{}.
\newblock \showarticletitle{An introduction to variational autoencoders}.
\newblock \bibinfo{journal}{\emph{Found. Trends Mach. Learn.}}
  \bibinfo{volume}{12}, \bibinfo{number}{4} (\bibinfo{year}{2019}),
  \bibinfo{pages}{307--392}.
\newblock


\bibitem[\protect\citeauthoryear{Kitamura, Pas, Lula, Lawton, and
  Benson}{Kitamura et~al\mbox{.}}{1996}]%
        {kitamura1996sequenced}
\bibfield{author}{\bibinfo{person}{Ryuichi Kitamura}, \bibinfo{person}{Eric~I
  Pas}, \bibinfo{person}{Clarisse~V Lula}, \bibinfo{person}{T~Keith Lawton},
  {and} \bibinfo{person}{Paul~E Benson}.} \bibinfo{year}{1996}\natexlab{}.
\newblock \showarticletitle{The sequenced activity mobility simulator (SAMS):
  an integrated approach to modeling transportation, land use and air quality}.
\newblock \bibinfo{journal}{\emph{Transportation}} \bibinfo{volume}{23},
  \bibinfo{number}{3} (\bibinfo{year}{1996}), \bibinfo{pages}{267--291}.
\newblock


\bibitem[\protect\citeauthoryear{Laub, Taimre, and Pollett}{Laub
  et~al\mbox{.}}{2015}]%
        {laub2015hawkes}
\bibfield{author}{\bibinfo{person}{Patrick~J Laub}, \bibinfo{person}{Thomas
  Taimre}, {and} \bibinfo{person}{Philip~K Pollett}.}
  \bibinfo{year}{2015}\natexlab{}.
\newblock \showarticletitle{Hawkes processes}.
\newblock \bibinfo{journal}{\emph{arXiv preprint arXiv:1507.02822}}
  (\bibinfo{year}{2015}).
\newblock


\bibitem[\protect\citeauthoryear{Liang, Ouyang, Yan, Wang, Tong, and
  Zimmermann}{Liang et~al\mbox{.}}{2021}]%
        {liang2021modeling}
\bibfield{author}{\bibinfo{person}{Yuxuan Liang}, \bibinfo{person}{Kun Ouyang},
  \bibinfo{person}{Hanshu Yan}, \bibinfo{person}{Yiwei Wang},
  \bibinfo{person}{Zekun Tong}, {and} \bibinfo{person}{Roger Zimmermann}.}
  \bibinfo{year}{2021}\natexlab{}.
\newblock \showarticletitle{Modeling Trajectories with Neural Ordinary
  Differential Equations.}. In \bibinfo{booktitle}{\emph{IJCAI}}.
  \bibinfo{pages}{1498--1504}.
\newblock


\bibitem[\protect\citeauthoryear{Limnios and Oprisan}{Limnios and
  Oprisan}{2012}]%
        {limnios2012semi}
\bibfield{author}{\bibinfo{person}{Nikolaos Limnios} {and}
  \bibinfo{person}{Gheorghe Oprisan}.} \bibinfo{year}{2012}\natexlab{}.
\newblock \bibinfo{booktitle}{\emph{Semi-Markov processes and reliability}}.
\newblock


\bibitem[\protect\citeauthoryear{Lin}{Lin}{1991}]%
        {lin1991divergence}
\bibfield{author}{\bibinfo{person}{Jianhua Lin}.}
  \bibinfo{year}{1991}\natexlab{}.
\newblock \showarticletitle{Divergence measures based on the Shannon entropy}.
\newblock \bibinfo{journal}{\emph{IEEE Transactions on Information theory}}
  \bibinfo{volume}{37}, \bibinfo{number}{1} (\bibinfo{year}{1991}),
  \bibinfo{pages}{145--151}.
\newblock


\bibitem[\protect\citeauthoryear{Macal and North}{Macal and North}{2005}]%
        {macal2005tutorial}
\bibfield{author}{\bibinfo{person}{Charles~M Macal} {and}
  \bibinfo{person}{Michael~J North}.} \bibinfo{year}{2005}\natexlab{}.
\newblock \showarticletitle{Tutorial on agent-based modeling and simulation}.
  In \bibinfo{booktitle}{\emph{WSC}}. IEEE, \bibinfo{pages}{14--pp}.
\newblock


\bibitem[\protect\citeauthoryear{Maslow}{Maslow}{1943}]%
        {maslow1943theory}
\bibfield{author}{\bibinfo{person}{Abraham~Harold Maslow}.}
  \bibinfo{year}{1943}\natexlab{}.
\newblock \showarticletitle{A theory of human motivation.}
\newblock \bibinfo{journal}{\emph{Psychological review}} \bibinfo{volume}{50},
  \bibinfo{number}{4} (\bibinfo{year}{1943}), \bibinfo{pages}{370}.
\newblock


\bibitem[\protect\citeauthoryear{Mei and Eisner}{Mei and Eisner}{2016}]%
        {mei2016neural}
\bibfield{author}{\bibinfo{person}{Hongyuan Mei} {and} \bibinfo{person}{Jason
  Eisner}.} \bibinfo{year}{2016}\natexlab{}.
\newblock \showarticletitle{The neural hawkes process: A neurally
  self-modulating multivariate point process}.
\newblock \bibinfo{journal}{\emph{arXiv preprint arXiv:1612.09328}}
  (\bibinfo{year}{2016}).
\newblock


\bibitem[\protect\citeauthoryear{Minor, Doppa, and Cook}{Minor
  et~al\mbox{.}}{2015}]%
        {minor2015data}
\bibfield{author}{\bibinfo{person}{Bryan Minor}, \bibinfo{person}{Janardhan~Rao
  Doppa}, {and} \bibinfo{person}{Diane~J Cook}.}
  \bibinfo{year}{2015}\natexlab{}.
\newblock \showarticletitle{Data-driven activity prediction: Algorithms,
  evaluation methodology, and applications}. In
  \bibinfo{booktitle}{\emph{KDD}}. \bibinfo{pages}{805--814}.
\newblock


\bibitem[\protect\citeauthoryear{Muri{\'c}, Tregubov, Blythe, Abeliuk,
  Choudhary, Lerman, and Ferrara}{Muri{\'c} et~al\mbox{.}}{2020}]%
        {muric2020massive}
\bibfield{author}{\bibinfo{person}{Goran Muri{\'c}}, \bibinfo{person}{Alexey
  Tregubov}, \bibinfo{person}{Jim Blythe}, \bibinfo{person}{Andr{\'e}s
  Abeliuk}, \bibinfo{person}{Divya Choudhary}, \bibinfo{person}{Kristina
  Lerman}, {and} \bibinfo{person}{Emilio Ferrara}.}
  \bibinfo{year}{2020}\natexlab{}.
\newblock \showarticletitle{Massive Cross-Platform Simulations of Online Social
  Networks}. In \bibinfo{booktitle}{\emph{AAMAS}}. \bibinfo{pages}{895--903}.
\newblock


\bibitem[\protect\citeauthoryear{North, Macal, Aubin, Thimmapuram, Bragen,
  Hahn, Karr, Brigham, Lacy, and Hampton}{North et~al\mbox{.}}{2010}]%
        {north2010multiscale}
\bibfield{author}{\bibinfo{person}{Michael~J North}, \bibinfo{person}{Charles~M
  Macal}, \bibinfo{person}{James~St Aubin}, \bibinfo{person}{Prakash
  Thimmapuram}, \bibinfo{person}{Mark Bragen}, \bibinfo{person}{June Hahn},
  \bibinfo{person}{James Karr}, \bibinfo{person}{Nancy Brigham},
  \bibinfo{person}{Mark~E Lacy}, {and} \bibinfo{person}{Delaine Hampton}.}
  \bibinfo{year}{2010}\natexlab{}.
\newblock \showarticletitle{Multiscale agent-based consumer market modeling}.
\newblock \bibinfo{journal}{\emph{Complexity}} \bibinfo{volume}{15},
  \bibinfo{number}{5} (\bibinfo{year}{2010}), \bibinfo{pages}{37--47}.
\newblock


\bibitem[\protect\citeauthoryear{Nweke, Teh, Al-Garadi, and Alo}{Nweke
  et~al\mbox{.}}{2018}]%
        {nweke2018deep}
\bibfield{author}{\bibinfo{person}{Henry~Friday Nweke},
  \bibinfo{person}{Ying~Wah Teh}, \bibinfo{person}{Mohammed~Ali Al-Garadi},
  {and} \bibinfo{person}{Uzoma~Rita Alo}.} \bibinfo{year}{2018}\natexlab{}.
\newblock \showarticletitle{Deep learning algorithms for human activity
  recognition using mobile and wearable sensor networks: State of the art and
  research challenges}.
\newblock \bibinfo{journal}{\emph{Expert Systems with Applications}}
  \bibinfo{volume}{105} (\bibinfo{year}{2018}), \bibinfo{pages}{233--261}.
\newblock


\bibitem[\protect\citeauthoryear{Ouyang, Shokri, Rosenblum, and Yang}{Ouyang
  et~al\mbox{.}}{2018}]%
        {ouyang2018non}
\bibfield{author}{\bibinfo{person}{Kun Ouyang}, \bibinfo{person}{Reza Shokri},
  \bibinfo{person}{David~S Rosenblum}, {and} \bibinfo{person}{Wenzhuo Yang}.}
  \bibinfo{year}{2018}\natexlab{}.
\newblock \showarticletitle{A Non-Parametric Generative Model for Human
  Trajectories.}. In \bibinfo{booktitle}{\emph{IJCAI}}.
  \bibinfo{pages}{3812--3817}.
\newblock


\bibitem[\protect\citeauthoryear{Pan, Huang, Li, Zhou, and Luo}{Pan
  et~al\mbox{.}}{2020}]%
        {pan2020xgail}
\bibfield{author}{\bibinfo{person}{Menghai Pan}, \bibinfo{person}{Weixiao
  Huang}, \bibinfo{person}{Yanhua Li}, \bibinfo{person}{Xun Zhou}, {and}
  \bibinfo{person}{Jun Luo}.} \bibinfo{year}{2020}\natexlab{}.
\newblock \showarticletitle{xGAIL: Explainable Generative Adversarial Imitation
  Learning for Explainable Human Decision Analysis}. In
  \bibinfo{booktitle}{\emph{KDD}}. \bibinfo{pages}{1334--1343}.
\newblock


\bibitem[\protect\citeauthoryear{Pfoser and Wenk}{Pfoser and Wenk}{[n.\,d.]}]%
        {pfosertowards}
\bibfield{author}{\bibinfo{person}{Andreas~Z{\"u}fle Pfoser} {and}
  \bibinfo{person}{Carola Wenk}.} \bibinfo{year}{[n.\,d.]}\natexlab{}.
\newblock \showarticletitle{Towards Large-Scale Agent-Based Geospatial
  Simulation}.
\newblock  (\bibinfo{year}{[n.\,d.]}).
\newblock


\bibitem[\protect\citeauthoryear{Prinz, Wu, Sarich, Keller, Senne, Held,
  Chodera, Sch{\"u}tte, and No{\'e}}{Prinz et~al\mbox{.}}{2011}]%
        {prinz2011markov}
\bibfield{author}{\bibinfo{person}{Jan-Hendrik Prinz}, \bibinfo{person}{Hao
  Wu}, \bibinfo{person}{Marco Sarich}, \bibinfo{person}{Bettina Keller},
  \bibinfo{person}{Martin Senne}, \bibinfo{person}{Martin Held},
  \bibinfo{person}{John~D Chodera}, \bibinfo{person}{Christof Sch{\"u}tte},
  {and} \bibinfo{person}{Frank No{\'e}}.} \bibinfo{year}{2011}\natexlab{}.
\newblock \showarticletitle{Markov models of molecular kinetics: Generation and
  validation}.
\newblock \bibinfo{journal}{\emph{The Journal of chemical physics}}
  \bibinfo{volume}{134}, \bibinfo{number}{17} (\bibinfo{year}{2011}),
  \bibinfo{pages}{174105}.
\newblock


\bibitem[\protect\citeauthoryear{Rabiner and Juang}{Rabiner and Juang}{1986}]%
        {rabiner1986introduction}
\bibfield{author}{\bibinfo{person}{Lawrence Rabiner} {and}
  \bibinfo{person}{Biinghwang Juang}.} \bibinfo{year}{1986}\natexlab{}.
\newblock \showarticletitle{An introduction to hidden Markov models}.
\newblock \bibinfo{journal}{\emph{ieee assp magazine}} \bibinfo{volume}{3},
  \bibinfo{number}{1} (\bibinfo{year}{1986}), \bibinfo{pages}{4--16}.
\newblock


\bibitem[\protect\citeauthoryear{Rauschenberger, Schmitt, and
  Hunter}{Rauschenberger et~al\mbox{.}}{1980}]%
        {rauschenberger1980test}
\bibfield{author}{\bibinfo{person}{John Rauschenberger}, \bibinfo{person}{Neal
  Schmitt}, {and} \bibinfo{person}{John~E Hunter}.}
  \bibinfo{year}{1980}\natexlab{}.
\newblock \showarticletitle{A test of the need hierarchy concept by a Markov
  model of change in need strength}.
\newblock \bibinfo{journal}{\emph{Administrative Science Quarterly}}
  (\bibinfo{year}{1980}), \bibinfo{pages}{654--670}.
\newblock


\bibitem[\protect\citeauthoryear{Recker and Root}{Recker and Root}{1981}]%
        {recker1981toward}
\bibfield{author}{\bibinfo{person}{WW Recker} {and} \bibinfo{person}{GS Root}.}
  \bibinfo{year}{1981}\natexlab{}.
\newblock \showarticletitle{Toward a dynamic model of individual activity
  pattern formulation}.
\newblock  (\bibinfo{year}{1981}).
\newblock


\bibitem[\protect\citeauthoryear{Shi, Yu, Da, Chen, and Zeng}{Shi
  et~al\mbox{.}}{2019}]%
        {shi2019virtual}
\bibfield{author}{\bibinfo{person}{Jing-Cheng Shi}, \bibinfo{person}{Yang Yu},
  \bibinfo{person}{Qing Da}, \bibinfo{person}{Shi-Yong Chen}, {and}
  \bibinfo{person}{An-Xiang Zeng}.} \bibinfo{year}{2019}\natexlab{}.
\newblock \showarticletitle{Virtual-taobao: Virtualizing real-world online
  retail environment for reinforcement learning}. In
  \bibinfo{booktitle}{\emph{AAAI}}, Vol.~\bibinfo{volume}{33}.
  \bibinfo{pages}{4902--4909}.
\newblock


\bibitem[\protect\citeauthoryear{Stab and Gurevych}{Stab and Gurevych}{2014}]%
        {stab2014annotating}
\bibfield{author}{\bibinfo{person}{Christian Stab} {and} \bibinfo{person}{Iryna
  Gurevych}.} \bibinfo{year}{2014}\natexlab{}.
\newblock \showarticletitle{Annotating argument components and relations in
  persuasive essays}. In \bibinfo{booktitle}{\emph{ACL}}.
  \bibinfo{pages}{1501--1510}.
\newblock


\bibitem[\protect\citeauthoryear{Sutton, Barto, et~al\mbox{.}}{Sutton
  et~al\mbox{.}}{1998}]%
        {sutton1998introduction}
\bibfield{author}{\bibinfo{person}{Richard~S Sutton}, \bibinfo{person}{Andrew~G
  Barto}, {et~al\mbox{.}}} \bibinfo{year}{1998}\natexlab{}.
\newblock \bibinfo{booktitle}{\emph{Introduction to reinforcement learning}}.
  Vol.~\bibinfo{volume}{135}.
\newblock


\bibitem[\protect\citeauthoryear{Wang, Wu, and Chou}{Wang
  et~al\mbox{.}}{2010}]%
        {wang2010toward}
\bibfield{author}{\bibinfo{person}{Chen-Ya Wang}, \bibinfo{person}{Yueh-Hsun
  Wu}, {and} \bibinfo{person}{Seng-Cho~T Chou}.}
  \bibinfo{year}{2010}\natexlab{}.
\newblock \showarticletitle{Toward a ubiquitous personalized daily-life
  activity recommendation service with contextual information: a services
  science perspective}.
\newblock \bibinfo{journal}{\emph{INF SYST E-BUS MANAG}} \bibinfo{volume}{8},
  \bibinfo{number}{1} (\bibinfo{year}{2010}), \bibinfo{pages}{13--32}.
\newblock


\bibitem[\protect\citeauthoryear{Wei, Xu, Liang, and Li}{Wei
  et~al\mbox{.}}{2021}]%
        {wei2021we}
\bibfield{author}{\bibinfo{person}{Hua Wei}, \bibinfo{person}{Dongkuan Xu},
  \bibinfo{person}{Junjie Liang}, {and} \bibinfo{person}{Zhenhui Li}.}
  \bibinfo{year}{2021}\natexlab{}.
\newblock \showarticletitle{How Do We Move: Modeling Human Movement with System
  Dynamics}. In \bibinfo{booktitle}{\emph{AAAI}}.
\newblock


\bibitem[\protect\citeauthoryear{Wu, Chen, Sun, Zheng, and Wang}{Wu
  et~al\mbox{.}}{2017}]%
        {wu2017modeling}
\bibfield{author}{\bibinfo{person}{Hao Wu}, \bibinfo{person}{Ziyang Chen},
  \bibinfo{person}{Weiwei Sun}, \bibinfo{person}{Baihua Zheng}, {and}
  \bibinfo{person}{Wei Wang}.} \bibinfo{year}{2017}\natexlab{}.
\newblock \showarticletitle{Modeling trajectories with recurrent neural
  networks}. IJCAI.
\newblock


\bibitem[\protect\citeauthoryear{Yang, Zhang, Zheng, and Yu}{Yang
  et~al\mbox{.}}{2014}]%
        {yang2014modeling}
\bibfield{author}{\bibinfo{person}{Dingqi Yang}, \bibinfo{person}{Daqing
  Zhang}, \bibinfo{person}{Vincent~W Zheng}, {and} \bibinfo{person}{Zhiyong
  Yu}.} \bibinfo{year}{2014}\natexlab{}.
\newblock \showarticletitle{Modeling user activity preference by leveraging
  user spatial temporal characteristics in LBSNs}.
\newblock \bibinfo{journal}{\emph{TSMC}} \bibinfo{volume}{45},
  \bibinfo{number}{1} (\bibinfo{year}{2014}), \bibinfo{pages}{129--142}.
\newblock


\bibitem[\protect\citeauthoryear{Ye, Zhu, and Cheng}{Ye et~al\mbox{.}}{2013}]%
        {ye2013s}
\bibfield{author}{\bibinfo{person}{Jihang Ye}, \bibinfo{person}{Zhe Zhu}, {and}
  \bibinfo{person}{Hong Cheng}.} \bibinfo{year}{2013}\natexlab{}.
\newblock \showarticletitle{What's your next move: User activity prediction in
  location-based social networks}. In \bibinfo{booktitle}{\emph{Proceedings of
  the 2013 SIAM International Conference on Data Mining}}. SIAM,
  \bibinfo{pages}{171--179}.
\newblock


\bibitem[\protect\citeauthoryear{Yu, Zhang, Wang, and Yu}{Yu
  et~al\mbox{.}}{2017}]%
        {yu2017seqgan}
\bibfield{author}{\bibinfo{person}{Lantao Yu}, \bibinfo{person}{Weinan Zhang},
  \bibinfo{person}{Jun Wang}, {and} \bibinfo{person}{Yong Yu}.}
  \bibinfo{year}{2017}\natexlab{}.
\newblock \showarticletitle{Seqgan: Sequence generative adversarial nets with
  policy gradient}. In \bibinfo{booktitle}{\emph{AAAI}},
  Vol.~\bibinfo{volume}{31}.
\newblock


\bibitem[\protect\citeauthoryear{Yuan, Ding, Wang, Jin, and Li}{Yuan
  et~al\mbox{.}}{2022}]%
        {yuan2022activity}
\bibfield{author}{\bibinfo{person}{Yuan Yuan}, \bibinfo{person}{Jingtao Ding},
  \bibinfo{person}{Huandong Wang}, \bibinfo{person}{Depeng Jin}, {and}
  \bibinfo{person}{Yong Li}.} \bibinfo{year}{2022}\natexlab{}.
\newblock \showarticletitle{Activity Trajectory Generation via Modeling
  Spatiotemporal Dynamics}. In \bibinfo{booktitle}{\emph{KDD}}.
  \bibinfo{pages}{4752--4762}.
\newblock


\bibitem[\protect\citeauthoryear{Zhang, Li, Zhou, and Luo}{Zhang
  et~al\mbox{.}}{2019}]%
        {zhang2019unveiling}
\bibfield{author}{\bibinfo{person}{Xin Zhang}, \bibinfo{person}{Yanhua Li},
  \bibinfo{person}{Xun Zhou}, {and} \bibinfo{person}{Jun Luo}.}
  \bibinfo{year}{2019}\natexlab{}.
\newblock \showarticletitle{Unveiling taxi drivers' strategies via cgail:
  Conditional generative adversarial imitation learning}. In
  \bibinfo{booktitle}{\emph{ICDM}}. IEEE, \bibinfo{pages}{1480--1485}.
\newblock


\bibitem[\protect\citeauthoryear{Zuo, Jiang, Li, Zhao, and Zha}{Zuo
  et~al\mbox{.}}{2020}]%
        {zuo2020transformer}
\bibfield{author}{\bibinfo{person}{Simiao Zuo}, \bibinfo{person}{Haoming
  Jiang}, \bibinfo{person}{Zichong Li}, \bibinfo{person}{Tuo Zhao}, {and}
  \bibinfo{person}{Hongyuan Zha}.} \bibinfo{year}{2020}\natexlab{}.
\newblock \showarticletitle{Transformer hawkes process}. In
  \bibinfo{booktitle}{\emph{ICML}}. PMLR, \bibinfo{pages}{11692--11702}.
\newblock


\end{thebibliography}

\clearpage

\end{document}